\documentclass[manuscript,screen]{acmart}
\usepackage[utf8]{inputenc}

\AtBeginDocument{%
  \providecommand\BibTeX{{%
    Bib\TeX}}}

\setcopyright{acmlicensed}
\copyrightyear{2024}
\acmYear{2024}
\acmDOI{XXXXXXX.XXXXXXX}

\usepackage{amsmath,amsfonts}
\usepackage{algorithmic}
\usepackage{array}
\usepackage[caption=false,font=normalsize,labelfont=sf,textfont=sf]{subfig}
\usepackage{textcomp}
\usepackage{stfloats}
\usepackage{multirow}
\usepackage{microtype} 
\usepackage{url}
\usepackage{verbatim}
\usepackage{graphicx}
\def\BibTeX{{\rm B\kern-.05em{\sc i\kern-.025em b}\kern-.08em
    T\kern-.1667em\lower.7ex\hbox{E}\kern-.125emX}}
\usepackage{balance}
\usepackage{natbib} 
\usepackage{float}
\usepackage{tikz}

\usepackage{amssymb}
\usepackage{pifont}

\usepackage{longtable}

\usepackage[edges]{forest}
\definecolor{hidden-draw}{RGB}{205, 44, 36}
\definecolor{hidden-blue}{RGB}{194,232,247}
\definecolor{hidden-orange}{RGB}{243,202,120}
\definecolor{hidden-yellow}{RGB}{255,229,204}
\definecolor{hidden-red}{RGB}{255,204,204}
\definecolor{hidden-green}{RGB}{204,255,204}
\usepackage[figuresright]{rotating}

\usepackage[figuresright]{rotating}
\usepackage{graphicx}
\usepackage{multirow}

\usepackage{tabularx}

\usepackage{array} 

\usepackage{graphicx}
\usepackage{subcaption}
\usepackage{subcaption}

\usepackage{amssymb}

\begin{document}

\tikzstyle{my-box}=[
    rectangle,
    draw=hidden-draw,
    rounded corners,
    text opacity=1,
    minimum height=1.5em,
    minimum width=5em,
    inner sep=2pt,
    align=center,
    fill opacity=.5,
    line width=0.8pt,
]
\tikzstyle{leaf}=[
    my-box, 
    minimum height=1.5em,
    text=black, 
    align=left,
    font=\normalsize,
    inner xsep=2pt,
    inner ysep=4pt,
    line width=0.8pt,
    draw=none, 
]

\title{Out-of-Distribution Detection: A Task-Oriented Survey of Recent Advances}

\author{Shuo Lu}
\affiliation{
\institution{NLPR \& MAIS, Institute of Automation, Chinese Academy of Sciences; UCAS}
\city{Beijing}
\country{China}}

\author{Yingsheng Wang}
\affiliation{
\institution{Anhui University}
\city{Hefei}
\country{China}}
\authornote{The first two authors contributed equally to this research. Emails: shuolucs@gmail.com, wangys200923@gmail.com.}

\author{Lijun Sheng}
\affiliation{
\institution{University of Science and Technology of China}
\city{Hefei}
\country{China}}

\author{Lingxiao He}
\affiliation{
\institution{Meituan}
\city{Beijing}
\country{China}}

\author{Aihua Zheng}
\affiliation{
\institution{Anhui University}
\city{Hefei}
\country{China}}

\author{Jian Liang}
\affiliation{
\institution{NLPR \& MAIS, Institute of Automation, Chinese Academy of Sciences; UCAS}
\city{Beijing}
\country{China}}
\authornote{Corresponding author. Email: liangjian92@gmail.com.}
\renewcommand{\shortauthors}{Lu et al.}

\begin{abstract}
Out-of-distribution (OOD) detection aims to detect test samples outside the training category space, which is an essential component in building reliable machine learning systems.
Existing reviews on OOD detection primarily focus on method taxonomy, surveying the field by categorizing various approaches. However, many recent works concentrate on non-traditional OOD detection scenarios, such as test-time adaptation, multi-modal data sources and other novel contexts. 
In this survey, we uniquely review recent advances in OOD detection from the task-oriented perspective for the first time.
According to the user’s access to the model, that is, whether the OOD detection method is allowed to modify or retrain the model, we classify the methods as training-driven or training-agnostic.
Besides, considering the rapid development of pre-trained models, large pre-trained model-based OOD detection is also regarded as an important category and discussed separately.
Furthermore, we provide a discussion of the evaluation scenarios, a variety of applications, and several future research directions.
We believe this survey with new taxonomy will benefit the proposal of new methods and the expansion of more practical scenarios.
A curated list of related papers is provided in the Github repository: \url{https://github.com/shuolucs/Awesome-Out-Of-Distribution-Detection}
\end{abstract}



\received{22 May 2024}
\received[revised]{19 June 2025}
\received[revised]{23 July 2025}
\received[accepted]{04 August 2025}

\maketitle

\section{Introduction}
Machine learning methods have made significant progress under the closed-world assumption, where test data is drawn from the same category as the training set, known as in-distribution (ID).
However, in the real world, models inevitably encounter test samples that do not belong to any training set category, commonly referred to as out-of-distribution (OOD) data.
OOD detection ~\citep{hendrycks2016baseline} aims to identify and reject OOD samples rather than make overconfident predictions arbitrarily ~\citep{bgf:HighConfWron_1} while maintaining accurate classification for ID data.
Models with superior OOD detection capabilities are more reliable and have important applications in numerous security-critical scenarios.
For instance, in medical diagnosis systems, a model that cannot detect OOD samples will misjudge unknown diseases and cause serious misdiagnosis ~\citep{schlegl2017unsupervised}.
Similarly, autonomous driving algorithms ~\citep{geiger2012we} should detect unknown scenarios and resort to human control to avoid accidents caused by arbitrary judgment.

\begin{table}[htbp]
\centering
\caption{Comparison of recent survey papers on OOD detection. ``Type'' indicates whether the survey is organized from a methodological perspective (``Method'') or from a broader task-oriented perspective (``Task''). ``OOD-Only'' indicates an exclusive focus on OOD detection; ``LPMs'' refers to coverage of methods based on large pretrained models; ``Application'' refers to in-depth analysis of real-world use cases; ``Multi-Modal'' indicates consideration of multiple data modalities; ``Test-Time'' indicates surveys that explicitly distinguish test-time adaptive methods from post-hoc approaches and provide dedicated discussion on test-time adaptation.}
\resizebox{0.8\textwidth}{!}{
\begin{tabular}{lccccccc}
\toprule
Study & Year & Type &   OOD-Only &  LPMs &  Application &  Multi-Modal &  Test-Time \\
\midrule
Salehi et al.~\cite{salehi2021unified} & 2021 & Method &\textcolor{red}{\ding{55}}&\textcolor{red}{\ding{55}}&\textcolor{red}{\ding{55}}&\textcolor{red}{\ding{55}}&\textcolor{red}{\ding{55}}\\
Yang et al.~\cite{yang2021generalized} & 2021 & Method &\textcolor{red}{\ding{55}}&\textcolor{red}{\ding{55}}&\textcolor{green}{\ding{51}}&\textcolor{red}{\ding{55}}&\textcolor{red}{\ding{55}}\\
Cui et al.~\cite{cui2022out} & 2022 & Method &\textcolor{green}{\ding{51}}&\textcolor{red}{\ding{55}}&\textcolor{green}{\ding{51}}&\textcolor{red}{\ding{55}}&\textcolor{red}{\ding{55}}\\
Lang et al.~\cite{lang2023survey_nlpood} & 2023 & Method &\textcolor{green}{\ding{51}}&\textcolor{red}{\ding{55}}&\textcolor{red}{\ding{55}}&\textcolor{red}{\ding{55}}&\textcolor{red}{\ding{55}}\\
Xu et al.~\cite{xu2024large} & 2024 & Method &\textcolor{red}{\ding{55}}&\textcolor{green}{\ding{51}}&\textcolor{red}{\ding{55}}&\textcolor{green}{\ding{51}}& \textcolor{red}{\ding{55}} \\
Miyai et al.~\cite{miyai2024generalized} & 2024 & Method &\textcolor{red}{\ding{55}}&\textcolor{green}{\ding{51}}&\textcolor{red}{\ding{55}}&\textcolor{red}{\ding{55}}&\textcolor{red}{\ding{55}}\\
Ours & 2024 & Task &\textcolor{green}{\ding{51}}&\textcolor{green}{\ding{51}}&\textcolor{green}{\ding{51}}&\textcolor{green}{\ding{51}}&\textcolor{green}{\ding{51}} \\
\bottomrule
\end{tabular}
}
\label{tab:survey-comparison}
\end{table}

Notably, several previous efforts have been dedicated to surveying and summarizing OOD detection in recent years.
~\citet{salehi2021unified} offer a  review covering Anomaly Detection, Novelty Detection, Open-Set Recognition and OOD Detection, and analyzes the relationships and distinctions among these fields.
~\citet{yang2021generalized} propose a unified framework to discuss OOD detection with several similar topics and categorize existing work into classification-based, density-based, distance-based and reconstruction-based methods.
Concurrent with our work, ~\citet{miyai2024generalized} discuss CLIP-based methods across related domains, while our survey specifically centers on OOD detection and provides a more comprehensive coverage of recent advances in this area.
~\citet{cui2022out} conduct a survey on OOD detection from a methodological perspective but with an alternative classification criterion, including supervised, semi-supervised, and unsupervised methods.
Additionally, ~\citet{lang2023survey_nlpood, xu2024large} review various OOD detection methods in the context of natural language processing.
However, previous works focus too much on the discussion from the perspective of methods and lack an in-depth exploration from the viewpoint of task scenarios.
Establishing a clear taxonomy of task scenarios can enhance a comprehensive understanding of the field and assist practitioners in selecting the appropriate method.
Moreover, given the recent introduction of new paradigms (e.g., test-time learning paradigm ~\citep{yang2023auto, geng2023soda, gao2023atta, du2024does, test-time_ETLT,wang2024goodat}) and methods based on large pre-trained models ~\citep{esmaeilpour2022zeroZOC,nguyen2024zero,fu2024clipscope,dai2023exploringLLM,liu2023good}, there is an urgent need for a comprehensive survey that incorporates the latest technologies.

\begin{figure*}[th!]
    \centering
    \resizebox{0.95\textwidth}{!}{
        \begin{forest}
            forked edges,
            for tree={
                grow=east,
                reversed=true,
                anchor=base west,
                parent anchor=east,
                child anchor=west,
                base=left,
                font=\large,
                rectangle,
                rounded corners,
                align=left,
                minimum width=4em,
                edge+={darkgray, line width=1pt},
                s sep=7pt,
                inner xsep=2pt,
                inner ysep=3pt,
                line width=0.8pt,
                ver/.style={rotate=90, child anchor=north, parent anchor=south, anchor=center},
            },
            where level=1{text width=14em,font=\normalsize,}{},
            where level=2{text width=13em,font=\normalsize,}{},
            where level=3{text width=11em,font=\normalsize,}{},
            where level=4{text width=10em,font=\normalsize,}{},
            [
                OOD Detection ,fill=hidden-red!70, draw=hidden-red!80!black,ver
                [
                    Problem: Training-driven  \\OOD detection(\S ~\ref{Training-driven OOD Detection}),fill=hidden-blue!70, draw=hidden-blue!80!black, align=center, text width=12em
                    [
                        Approaches with only \\ID Data  (\S ~\ref{only-id}),fill=hidden-blue!50, draw=hidden-blue!80!black,align=center
                            [
                            Reconstruction-based, fill=hidden-blue!30, draw=hidden-blue!80!black,
                                [
                                MoodCat ~\citep{yang2022out}{,}
                                RAE~\citep{zhou2022rethinking}{,}
                                MOOD ~\citep{li2023rethinking}{,}
                                MOODv2 ~\citep{li2024moodv2}{,}
                                PRE ~\citep{osada2023out}{,}
                                LMD ~\citep{liu2023unsupervised}
                                DiffGuard ~\citep{gao2023diffguard}{,}
                                \\DenoDiff~\citep{graham2023denoising}
                                , leaf, text width=38em,fill=hidden-blue!20, draw=hidden-blue!80!black,
                                ]
                            ]
                            [
                            Probability-based, fill=hidden-blue!30, draw=hidden-blue!80!black,
                                [
                                HVCM~\citep{li2023hierarchical}{,}
                                DDR~\citep{huang2022density}{,}
                                LID~\citep{kamkari2024geometric}
                                , leaf, text width=38em,fill=hidden-blue!20, draw=hidden-blue!80!black,
                                ]
                            ]
                            [
                            Logits-based, fill=hidden-blue!30, draw=hidden-blue!80!black,
                                [
                                LogitNorm ~\citep{wei2022mitigating}{,}
                                UE-NL ~\citep{huang2023uncertainty}{,}
                                DML ~\citep{zhang2023decoupling}
                                , leaf, text width=38em,fill=hidden-blue!20, draw=hidden-blue!80!black,
                                ]
                            ]
                            [
                            OOD Synthesis, fill=hidden-blue!30, draw=hidden-blue!80!black,
                                [
                                ConfiCali~\citep{lee2017training}{,}
                                CODEs ~\citep{tang2021codes}{,}
                                CMG ~\citep{wang2022cmg}{,}
                                VOS ~\citep{du2022vos}{,}
                                NPOS ~\citep{tao2023non}{,}
                                SHIFT ~\citep{kwon2023improving}{,}
                                ATOL ~\citep{zheng2023out}{,}
                                \\SSOD ~\citep{pei2023image}{,}
                                SEM ~\citep{yang2023full_food}{,}
                                Forte ~\citep{ganguly2024forte}{,}
                                HamOS ~\citep{li2025outlier}{,}
                                POP ~\citep{gong2025out}
                                , leaf, text width=38em,fill=hidden-blue!20, draw=hidden-blue!80!black,
                                ]
                            ]
                            [
                            Prototype-based, fill=hidden-blue!30, draw=hidden-blue!80!black,
                                [
                                Step ~\citep{zhou2021step}{,}
                                SIREN ~\citep{du2022siren}{,}
                                CIDER ~\citep{ming2022exploit}{,}
                                PALM ~\citep{lu2024learning}{,}
                                ReweightOOD ~\citep{regmi2024reweightood}{,}
                                AROS ~\citep{mirzaei2024adversarially}{,}
                                PFS ~\citep{wu2024pursuing}
                                , leaf, text width=38em,fill=hidden-blue!20, draw=hidden-blue!80!black,
                                ]
                            ]
                            [
                            A special case:\\Long-tail ID data, fill=hidden-blue!30, draw=hidden-blue!80!black,
                                [
                                OLTR ~\citep{liu2019large}{,}
                                II-Mixup~\citep{mehta2022out}{,}
                                AREO ~\citep{sapkota2022adaptive}{,}
                                IDCP~\citep{jiang2023detecting}{,}
                                Open-Sampling ~\citep{wei2022open}{,}
                                COOD ~\citep{hogeweg2024cood}
                                , leaf, text width=38em,fill=hidden-blue!20, draw=hidden-blue!80!black,
                                ]
                            ]
                    ]
                    [
                       Approaches with both ID \\ and OOD Data  (\S ~\ref{Approaches with Both ID and OOD Data}),fill=hidden-blue!50, draw=hidden-blue!80!black,align=center
                        [
                            Boundary Regularization,fill=hidden-blue!30, draw=hidden-blue!80!black,
                                [
                                OE ~\citep{hendrycks2018deep}{,}
                                ELOC~\citep{vyas2018out}{,}
                                Why-ReLU~\citep{hein2019relu}{,}
                                SSL-GOOD~\citep{mohseni2020self}{,}
                                EnergyOE ~\citep{liu2020energy}{,}
                                MixOE ~\citep{zhang2023mixture}
                                , leaf, text width=38em,fill=hidden-blue!20, draw=hidden-blue!80!black,
                                ]
                        ]
                        [
                            Outlier Mining,fill=hidden-blue!30, draw=hidden-blue!80!black,
                                [
                                BD-Resamp~\citep{li2020background}{,}
                                POEM ~\citep{ming2022poem}{,}
                                DAOL ~\citep{wang2023learning}{,}
                                DOE~\citep{wang2023out}{,}
                                MixOE ~\citep{zhang2023mixture}{,}
                                DivOE ~\citep{zhu2023diversified}
                                , leaf, text width=38em,fill=hidden-blue!20, draw=hidden-blue!80!black,
                                ]
                        ]
                        [
                            Imbalanced ID,fill=hidden-blue!30, draw=hidden-blue!80!black,
                            [
                                PASCAL ~\citep{wang2022partial}{,}
                                COCL ~\citep{miao2023out}{,}
                                BERL~\citep{choi2023balanced}{,}
                                EAT ~\citep{wei2023eat}
                                , leaf, text width=38em,fill=hidden-blue!20, draw=hidden-blue!80!black,
                            ]
                        ]
                    ]
                ]
                [
                    Problem: Training-agnostic \\OOD detection (\S ~\ref{sec:Training-agnostic OOD Detection}),fill=hidden-blue!70, draw=hidden-blue!80!black,align=center,text width=12em
                    [
                        Post-hoc Approaches (\S ~\ref{post-hoc}),fill=hidden-blue!50, draw=hidden-blue!80!black,align=center
                            [
                            Output-based,fill=hidden-blue!30, draw=hidden-blue!80!black,
                            [
                                MSP ~\citep{hendrycks2016baseline}{,}
                                MaxLogits ~\citep{hendrycks2019scaling}{,}
                                Energy ~\citep{liu2020energy}{,}
                                GEN ~\citep{liu2023gen}{,}
                                ZODE ~\citep{xue2024enhancing}{,}
                                LogicOOD ~\citep{kirchheim2024out}
                                , leaf, text width=38em,fill=hidden-blue!20, draw=hidden-blue!80!black,
                            ]
                        ]
                        [
                            Distance-based,fill=hidden-blue!30, draw=hidden-blue!80!black,
                            [
                                Mahalanobis ~\citep{lee2018simple}{,}
                                NNGuide ~\citep{park2023nearest}{,}
                                KNN ~\citep{sun2022out}{,}
                                SSD ~\citep{sehwag2021ssd}{,}Mahalanobis++~\citep{mueller2025mahalanobis}
                                , leaf, text width=38em,fill=hidden-blue!20, draw=hidden-blue!80!black,
                            ]
                        ]
                        [
                            Gradient-based,fill=hidden-blue!30, draw=hidden-blue!80!black,
                            [
                                Grad ~\citep{lee2020gradients}{,}
                                GradNorm ~\citep{huang2021importance}{,}
                                GradOrth ~\citep{behpour2023gradorth}{,}
                                GAIA ~\citep{chen2023gaia}{,}
                                OPNP ~\citep{chen2024optimal}{,}
                                PRO ~\citep{chen2025leveraging}
                                , leaf, text width=38em,fill=hidden-blue!20, draw=hidden-blue!80!black,
                            ]
                        ]
                        [
                            Feature-based,fill=hidden-blue!30, draw=hidden-blue!80!black,
                            [
                                ODIN ~\citep{liang2017enhancing}{,}
                                ReAct ~\citep{sun2021react}{,}
                                VRA ~\citep{xu2023vra}{,}
                                SHE ~\citep{zhang2022out}{,}
                                ViM ~\citep{wang2022vim}{,}
                                Neco ~\citep{ammar2023neco}{,}
                                ASH ~\citep{djurisic2022extremely}{,}
                                NAC ~\citep{liu2023neuron}
                                \\ Optimal-FS~\citep{zhao2024towards}{,}
                                BLOOD ~\citep{jelenic2023out}{,}
                                SCALE ~\citep{xu2023scaling}{,}
                                DDCS ~\citep{yuan2024discriminability}{,}
                                LINe ~\citep{yong2023line}{,}
                                KANs ~\citep{canevaro2025advancing}{,}
                                \\CADRef ~\citep{ling2025cadref}{,}
                                ITP ~\citep{xu2025itp}{,}
                                NCI ~\citep{liu2023detecting}
                                , leaf, text width=38em,fill=hidden-blue!20, draw=hidden-blue!80!black,
                            ]
                        ]
                        [
                            Density-based,fill=hidden-blue!30, draw=hidden-blue!80!black,
                            [
                                GEM ~\citep{morteza2022provable}{,}
                                ConjNorm ~\citep{peng2024conjnorm}
                                , leaf, text width=38em,fill=hidden-blue!20, draw=hidden-blue!80!black,
                            ]
                        ]
                    ]
                    [
                        Test-time Adaptive  Approaches \\(\S ~\ref{test-time}) ,fill=hidden-blue!50, draw=hidden-blue!80!black,align=center
                        [
                             Model-optimization-based,fill=hidden-blue!30, draw=hidden-blue!80!black,
                            [
                                WOODS ~\citep{woods}{, }
                                AUTO ~\citep{yang2023auto}{, }SODA ~\citep{geng2023soda}{, }ATTA ~\citep{gao2023atta}{, }SAL ~\citep{du2024does}
                                , leaf, text width=38em,fill=hidden-blue!20, draw=hidden-blue!80!black,
                            ]
                        ]
                        [
                             Model-optimization-free,fill=hidden-blue!30, draw=hidden-blue!80!black,
                            [
                                 ETLT ~\citep{test-time_ETLT}{, }AdaOOD ~\citep{AdaOOD}{, }GOODAT ~\citep{wang2024goodat}{, }RTL ~\citep{Fan2024CVPR}{, }OODD ~\citep{yang2025oodd}
                                , leaf, text width=38em,fill=hidden-blue!20, draw=hidden-blue!80!black,
                            ]
                        ]      
                    ]
                ]
                [
                    Problem: Large pre-trained \\model-based OOD detection\\(\S ~\ref{sec:Large Pre-trained Model based OOD Detection}),fill=hidden-green!70,draw=hidden-green!80!black,align=center,text width=12em
                    [
                        Zero-shot Approaches  \\(\S ~\ref{zero-shot-ood}),fill=hidden-green!50,draw=hidden-green!80!black,align=center
                            [
                             VLM-based,fill=hidden-green!20,draw=hidden-green!80!black,
                                [
                                CLIPScope ~\citep{fu2024clipscope}{, }
                                RONIN ~\citep{nguyen2024zero}{, }
                                ZOC ~\citep{esmaeilpour2022zeroZOC}{, }
                                MCM ~\citep{ming2022delvingMCM}{, }CLIPN ~\citep{wang2023clipn}{, }NegLabel ~\citep{jiang2023negative}{, }
                                \\NegPrompt-C~\citep{nie2023out}{, }
                                LAPT ~\citep{zhang2024lapt} {, } AdaNeg~\citep{zhang2024adaneg}{, } CSP~\citep{chen2024conjugated} {, } SimLabel~\citep{zou2025simlabel}  
                 {, } ~\citep{jung2024enhancing} {, }   \\SeTAR~\citep{li2024setar} {, }  OT-DETECTOR~\citep{liu2025ot} 
                                , leaf, text width=38em,fill=hidden-green!20,draw=hidden-green!80!black,
                                ]
                            ]
                            [
                             LLM-based,fill=hidden-green!20,draw=hidden-green!80!black,
                                [
                                WK-LLM~\citep{dai2023exploringLLM}{, }ODPC ~\citep{huang2024out}{, }LLM-OOD~\citep{liu2023good}{, } CMA~\citep{lee2025concept}{, }ReGuide~\citep{kim2024reflexive} {,} COOD~\citep{liu2024cood}{, }   \\EOE~\citep{cao2024envisioning}
                                , leaf, text width=38em,fill=hidden-green!20,draw=hidden-green!80!black,
                                ]
                            ]      
                    ]
                    [
                        Few-shot Approaches \\(\S ~\ref{few-shot-ood}),fill=hidden-green!50,draw=hidden-green!80!black,align=center
                            [
                            Fine-tuning based, fill=hidden-green!20,draw=hidden-green!80!black,
                                [
                                CLIP-OS ~\citep{sun2024clip}{,}
                                Local-Prompt ~\citep{zeng2025local}{,}
                                GalLop ~\citep{lafon2024gallop}{,}
                                ID-like ~\citep{bai2023id}{, } DSFG ~\citep{dong2023towardsDSFG}{, }LoCoOp ~\citep{miyai2023locoop}{, }\\NegPrompt ~\citep{li2024learning}{, }
                                SUPREME~\citep{wang2025mitigating}  {, }
                                SCT~\citep{yu2024self}  {, } GaCoOp~\citep{tong2025enhancing}
                                , leaf, text width=38em,fill=hidden-green!20,draw=hidden-green!80!black,
                                ]
                            ]
                            [
                            Meta-learning based, fill=hidden-green!20,draw=hidden-green!80!black,,
                                [
                                OOD-MAML ~\citep{jeong2020ood}{, }HyperMix ~\citep{mehta2024hypermix}
                                , leaf, text width=38em,fill=hidden-green!20,draw=hidden-green!80!black,
                                ]
                            ]
                    ]
                    [
                        Full-shot Approaches  \\(\S ~\ref{full-shot-ood}),fill=hidden-green!50,draw=hidden-green!80!black,align=center
                            [
                                NPOS ~\citep{tao2023non}{, } PT-OOD~\citep{miyai2023can}{, }
                                TOE ~\citep{park2023powerfulness}
                                , leaf, text width=48.5em,fill=hidden-green!20,draw=hidden-green!80!black,,text width=50.6em
                            ]
                    ]
                ]
            ]
        \end{forest}
    }
    \caption{
    Taxonomy of OOD detection problem scenarios and solutions. 
    The first two categories (in \textcolor{hidden-blue!90!black}{blue}) correspond to shallow model-based methods, while the third category (in \textcolor{hidden-green!70!black}{green}) highlights approaches based on foundation models (large pre-trained models). 
    Different colors are used to visually distinguish the traditional shallow-model-based solutions from the newly emerging foundation model-based methods, emphasizing the conceptual differences between them.
    }
    \label{fig:survey}
\end{figure*}
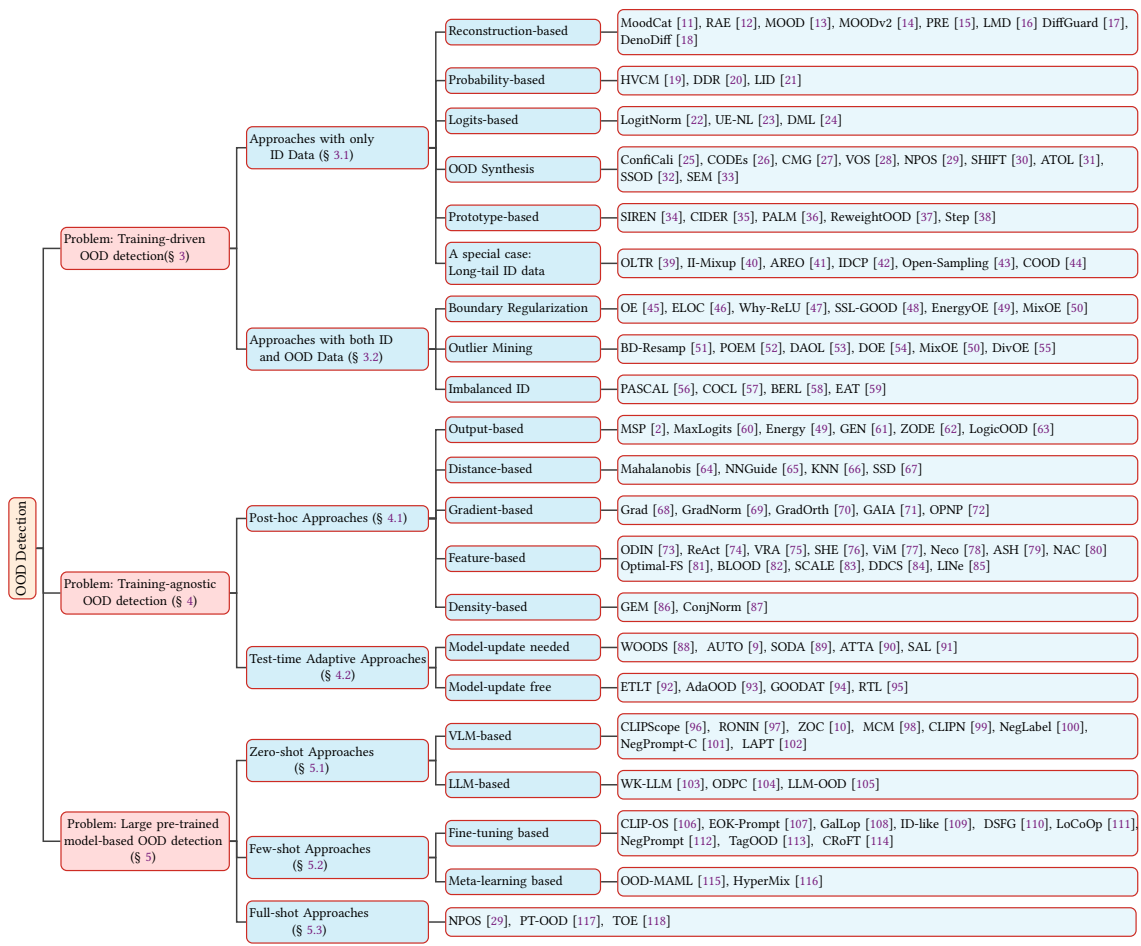

In this survey, we for the first time review the recent advances in OOD detection with a task-oriented taxonomy, as illustrated in Fig.~\ref{fig:survey}.
Based on whether the method needs to control the pre-training process, 
we categorize OOD detection algorithms into \textit{training-driven} and \textit{training-agnostic} methods.
Considering the rapid development of large pre-trained models nowadays and the evolving definition of OOD under such models, we also regard \textit{large pre-trained model-based} OOD detection as a separate section.
In detail, training-driven methods achieve high detection capability by designing the optimization process of the training stage.
They are further classified and discussed according to whether OOD data is used in training.
Training-agnostic methods distinguish OOD data from ID ones based on a well-trained model,
skipping the time-consuming and expensive pre-training process in practice.
According to whether utilizing test samples to further improve OOD detection performance, we categorize them into post-hoc and test-time methods.
Large pre-trained model-based OOD detection methods focus on models such as vision language models or large language models, which are pre-trained on vast datasets and excel in a wide array of tasks.
We discuss them in terms of whether they have access to a few examples, including zero-shot, few-shot and full-shot scenarios.

The structure of this survey is as follows. 
We discuss the related topics of OOD detection in Sec.~\ref{sec:related_work}. Next, we summarize training-driven OOD detection approaches in Sec.~\ref{Training-driven OOD Detection}, and introduce the training-agnostic OOD detection methods in Sec.~\ref{sec:Training-agnostic OOD Detection}. 
Then, in Sec.~\ref{sec:Large Pre-trained Model based OOD Detection}, we introduce large pre-trained model-based OOD detection. 
An overview of the evaluation metrics, experimental protocols, and applications is presented in Sec.~\ref{evaluationandapplication}. 
Following that, 
we discuss promising trends and open challenges in Sec.~\ref{emerging future work} to shed light on underexplored and potentially critical avenues.


\section{Related work}\label{sec:related_work}
\textbf{Anomaly Detection.} 
Anomaly detection (AD) identifies data points that deviate from normal patterns~\citep{chandola2009anomaly}, with wide applications in fraud detection~\citep{pourhabibi2020fraud}, network security~\citep{lazarevic2003comparative}, and system monitoring~\citep{du2017deeplog}.
While both AD and OOD detection identify unusual samples, AD differs in three main aspects. First, AD is typically unsupervised, using only normal samples for training, whereas OOD detection involves labeled categories. Second, AD focuses on perceptual anomalies, while OOD detects semantic shifts from unseen categories. Lastly, although both use AUROC~\citep{wehler2001short} for evaluation, AD employs AUPRO~\citep{bertoldo2024aupimo} for pixel-level performance.
Beyond these standard metrics, practical evaluation also assesses human-centric and operational criteria, such as \textbf{detection latency}~\citep{lavin2015evaluating,sinha2024real}, the economic \textbf{cost of errors}~\citep{xu2013cost,kim2022study}, model \textbf{interpretability}~\citep{jiang2023interpretability,doshi2023towards}, and \textbf{computational footprint}~\citep{ibidunmoye2015performance,huertas2023comparative}.

\textbf{Novelty Detection.} Novelty detection (ND) aims to identify previously unseen data points~\citep{pimentel2014review}, enabling systems to adapt to new conditions or scenarios. Unlike AD, which finds deviations within known patterns, ND discovers new patterns, such as emerging social media trends~\citep{li2017real}, novel species~\citep{pimentel2014review}, or new document topics~\citep{aksoy2012novelty}. In essence, ND detects surprises within familiar contexts, while OOD detection addresses data from entirely unfamiliar contexts.

\textbf{Open Set Recognition.} Open Set Recognition (OSR) extends traditional classification by identifying both known classes and instances from unknown categories~\citep{bendale2016towards}. Unlike OOD detection which identifies any OOD data, OSR specifically focuses on recognizing novel classes within the same domain. This makes it particularly relevant for applications like robotics~\citep{marsland2005line} and autonomous systems~\citep{amini2018variational}.

\textbf{Outlier Detection.} Outlier detection (OD) identifies individual data points that significantly deviate from the majority~\citep{singh2012outlier}. Unlike OOD detection which encounters outliers only during deployment, OD is transductive with inherent access to outliers. It finds applications in fraud detection and data cleaning~\citep{pawar2014survey}.

\textbf{Zero-shot learning.} Zero-shot learning aims to recognize objects without seeing their examples during training by transferring knowledge from seen to unseen classes~\citep{romera2015embarrassingly, wang2019survey}. Unlike OOD detection which flags anomalous data~\citep{lee2018simple}, zero-shot learning actively classifies new categories through semantic relationships~\citep{xian2018zero}.

\textbf{Selective Classification.} Selective classification (or reject option classification) allows models to abstain from predictions when confidence is insufficient ~\citep{geifman2017selective, chang2024survey}. While OOD detection identifies samples from different distributions, selective classification focuses on managing prediction uncertainty within the training distribution ~\citep{el2010foundations}.

\textbf{Misclassification (Failure) Detection.} Misclassification detection identifies incorrect predictions on ID data~\citep{ahmed2022failure, dong2025trustvlm, inceoglu2021fino,liu2024typicalness}, critical for domains like autonomous driving. It differs from OOD detection by focusing on errors within the training distribution~\citep{jaeger2022call}, rather than external inputs.

\vspace{-5pt}
\section{Problem: Training-driven OOD Detection}\label{Training-driven OOD Detection}

\begin{figure}
    \centering
    \includegraphics[width=0.8\textwidth]{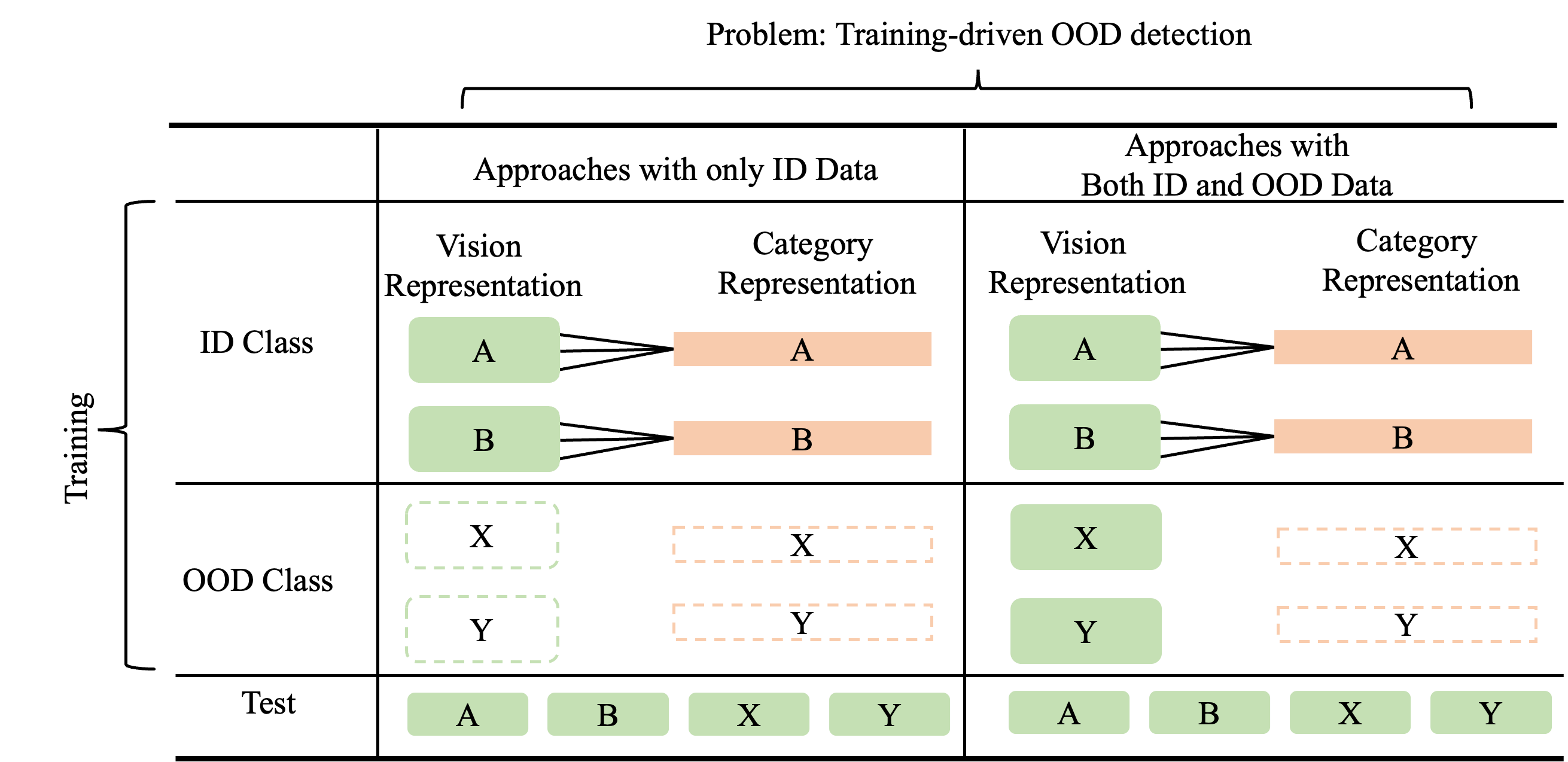}
    \caption{Illustration of training-driven OOD detection approaches. Dashed borders indicate that they are not used in the specific phase. OOD images are excluded in the ``Approaches with only ID data'' on the left but are included in the ``Approaches with both ID and OOD data'' on the right. In both cases, OOD labels are not utilized.}
    \label{fig:training-driven}
\end{figure}

In the \textit{training-driven OOD detection} problem, researchers design the pre-training process to obtain models with superior OOD detection capabilities.
Based on whether OOD data is accessible during training, we further divide methods under this scenario into two folds: 
training with only ID data and training with both ID and OOD data, as shown in Fig.~\ref{fig:training-driven}.
\vspace{-2pt}
\subsection{OOD Detection Approaches with only ID Data}\label{only-id}
\textbf{Overview.} 
Given the ID data, approaches in this section train a model on them and aim to utilize the model for detecting OOD test samples while ensuring classification performance of the ID data.
They focus specifically on mining information from ID data, without explicitly relying on other information from real OOD data.
We further differentiate these methods into the following five categories: \textit{Reconstruction-based}, \textit{Probability-based}, \textit{Logits-based}, \textit{OOD Synthesis}, and \textit{Prototype-based}. Considering real-world requirements, we also delve into a specific scenario: \textit{Long-tail ID data}.

\textbf{Reconstruction-based.} 
Reconstruction-based OOD detection methods focus on measuring representational differences between original and reconstructed data. These approaches leverage the model’s capability to recover input semantics, operating on the premise that OOD samples contain distinct semantic features from ID data, as shown in Fig.~\ref{fig:Training}~(a). Several notable works have advanced this direction. MoodCat ~\citep{yang2022out} employs masked image modeling with constrained synthesis based on classification outputs. ~\citet{zhou2022rethinking} enhances detection by incorporating feature activation analysis through an auxiliary module. Building upon masked image modeling, MOOD ~\citep{li2023rethinking} and its successor MOODv2 ~\citep{li2024moodv2} demonstrate improved capabilities in learning data distributions. PRE ~\citep{osada2023out} further develops this approach by combining normalized flow with typicality-based penalties to effectively distinguish between OOD and ID samples.

\begin{figure}
    \centering 
    \includegraphics[trim=1bp 1bp 1bp 1bp, clip, width=0.9\linewidth]{images/Training.png}
    \caption{Illustration of main Training-driven OOD detection approaches:
(a) Reconstruction-based: Identifies OOD samples by comparing the dissimilarity between original feature representation F(x) and its reconstruction F’(x) through networks like VAE or DDPM.
(b) Prototype-based: Learns optimal ID prototypes during training for OOD distinction, where h represents penultimate layer features.
(c) OOD Synthesis: Generates synthetic OOD samples by extrapolating from ID data when real OOD data is unavailable.
(d) Outlier Mining: Leverages available real OOD data to generate typical OOD samples during training.
(e) Boundary Regularization: Optimizes model decision boundary using real OOD data without synthesis.}
    \label{fig:Training}
\end{figure}

Recent advances in diffusion models~\citep{ho2020denoising} have led to improved training stability and image quality, prompting their application in OOD detection. \citet{graham2023denoising} introduce DDPM for reconstructing noise-corrupted images, using multidimensional reconstruction error for OOD identification and allowing external regulation of the information bottleneck. LMD~\citep{liu2023unsupervised} also applies diffusion models, disrupting and reconstructing data to separate OOD samples from the original manifold. DiffGuard~\citep{gao2023diffguard} leverages pre-trained diffusion models to enhance semantic discrepancies between reconstructed OOD and original images.

\textbf{Probability-based}. 
Probability-based approaches aim to establish probability models to describe the distribution of training data. 
In this domain, ~\citet{li2023hierarchical} model each ID category during training using multiple Gaussian mixture models~\citep{rasmussen1999infinite}, and during the prediction phase, it combines Mahalanobis distance metrics to assess the likelihood of anomalous classes. 
~\citet{huang2022density} address this issue by introducing two regularization constraints. 
The density consistency regularization aligns the analytical density with low-dimensional class labels, and the contrastive distribution regularization helps separate the density between ID and OOD samples.
Furthermore, LID \citep{kamkari2024geometric} introduces a new detection criterion for the OOD detection paradox in the context of data generation by deep generative models.
It measures whether data should be classified as ID by estimating the local intrinsic dimension (LID) of the learned manifold of the generative model, when the data is assigned a high probability and the probability mass is non-negligible.



\textbf{Logits-based.} 
These methods detect OOD samples by analyzing the logits (raw outputs before softmax) from the neural network’s final layer.
Logits typically represent the model's confidence or probability for each category.
LogitNorm ~\citep{wei2022mitigating} uses logit normalization to enforce a constant vector norm on logits during training to mitigate issues of model overconfidence.
UE-NL~\citep{huang2023uncertainty} leverages Bayesian principles to jointly learn embeddings and uncertainty scores, normalizing logits to improve robustness in OOD detection. DML~\citep{zhang2023decoupling} further advances this area by decoupling logits and balancing key components, thereby mitigating attribute interference and enhancing detection performance.

\textbf{OOD Synthesis.} In the task of OOD detection, incorporating the features of OOD data during model training can enhance the OOD detection performance by allowing the model to better recognize and differentiate OOD data.
Due to the challenges in acquiring distribution information for OOD samples, some methods employ ID data to estimate the distribution of OOD data, 
simulating real-world scenarios where a model encounters OOD data, as shown in Fig.~\ref{fig:Training} (c).

~\citet{lee2017training} and VOS~\citep{du2022vos} both synthesize OOD samples from low-density or low-likelihood regions of the ID feature space, but VOS enables adaptive outlier generation without external data or manual tuning. In contrast, NPOS~\citep{tao2023non} avoids distributional assumptions by using non-parametric density estimation to select boundary samples, enhancing flexibility and generality. SSOD~\citep{pei2023image} further differs by leveraging self-supervised sampling to extract OOD signals directly from ID backgrounds, addressing synthesis bias.
OOD detection methods often generate synthetic outliers from in-distribution data. Techniques include recombining features with GANs (CODEs~\citep{tang2021codes}), using conditional VAEs (CMG~\citep{wang2022cmg}), or applying Mixup (SEM~\citep{yang2023full_food}). Alternatively, other approaches estimate data typicality by analyzing the local geometric and density statistics of the feature manifold (FORTE~\citep{ganguly2024forte}).
SHIFT~\citep{kwon2023improving} leverages CLIP and latent diffusion models to generate OOD images by replacing ID object regions with contextually consistent features, though maintaining high sample quality remains a challenge. ATOL~\citep{zheng2023out} introduces an auxiliary task that constructs non-overlapping regions for ID and OOD data in latent space, using generated samples to reinforce this separation and aligning genuine ID data to enhance reliability.
Recent studies show that HamOS~\citep{li2025outlier} uses Hamiltonian Monte Carlo~\citep{duane1987hybrid} to generate virtual OOD anomalies by sampling from a Markov chain~\citep{neal1993probabilistic} when real OOD data is unavailable. It projects features onto a hypersphere and leverages K-nearest neighbor distances to guide anomaly generation. POP~\citep{gong2025out} instead refines decision boundaries using virtual OOD prototypes and penalizes misclassifications, enhancing OOD detection without explicit outlier synthesis.

\textbf{Prototype-based.} During the model training process, prototype-based OOD detection methods aim to model the ID data using prototypes to learn the common distribution characteristics of the ID data. 
For OOD data, there is a significant difference between the sample features and the ID prototype. In the testing phase, the model determines the sample's category by measuring the difference between the sample and the ID prototype.
The general procedure of these methods is outlined in Fig.~\ref{fig:Training} (b).


SIREN ~\citep{du2022siren} first models the distribution of ID data using the von Mises-Fisher(vMF)~\citep{mardia2009directional}  model, which enables representing each class as a compact cluster, i.e., a class prototype. 
Generally, the vMF distribution modeling formula can be represented as: $ p_{D}(z;p_{k},k) = Z_{D}(k)exp(kp_{k}^{T}z)$, where $p_{k}$ is the $k$-th prototype with unit norm, \(\kappa \geq 0\) represents the concentration around the mean, and \(Z_D(\kappa)\) is the normalization factor.
In the prototype-based approach, an embedding vector \( z \) is assigned to class \( c \) with the following normalized probability:
\begin{equation}
    p(y=c|z;(P_{j}, k_{j})_{j=1}^{C})=\frac{Z_{D}(k_{c})exp(k_{c}\mu_{c}^{T}z)}{{\textstyle \sum_{j=1}^{C}Z_{d}(k_{j})exp(k_{j}\mu_{j}^{T}z)}}, 
\end{equation}
where $c\in {\{1,2,...C\}}$.
Additionally, within the loss function, it enforces alignment between the embedding vectors of ID samples and class prototypes, constraining each ID class sample. 
This parametric OOD score can be directly obtained after training, without requiring separate estimation.
CIDER ~\citep{ming2022exploit} builds upon SIREN by jointly optimizing two losses to enhance data discriminability, encouraging the maximization of angular distances between prototypes of different classes and the internal compactness of prototypes within the same class.
The optimization process during model training pertains to the prototypes of classes.
AROS~\citep{mirzaei2024adversarially} enhances feature separation between ID and OOD data by applying the Lyapunov stability theorem, guiding embeddings of ID and OOD samples toward distinct stable points. Pseudo-OOD samples are generated by sampling from low-probability regions of the ID distribution.
CNC~\citep{harun2025controlling} controls neural collapse (NC) at different network stages: entropy regularization in the encoder alleviates NC, while the projector head promotes NC, resulting in more compact prototypes and improved OOD detection.
ReweightOOD ~\citep{regmi2024reweightood} argues that optimizing non-class data impedes achieving clear class separability, while focusing on fewer class data makes it challenging to achieve lower MSE scores.
To address this, they propose a re-weighting optimization strategy to balance the significance of different losses.
Although the ideas behind Step ~\citep{zhou2021step} are different, in the context of semi-supervised tasks, it essentially generates clusters of unlabeled ID and OOD samples through a contrastive learning process, which is conceptually similar to prototype learning.
However, PALM~\citep{lu2024learning} notes that representing each class with a single prototype often fails to capture the internal diversity of data. To address this, PALM adopts a mixed-prototype strategy, assigning multiple prototypes per class to better model informative representations. By jointly learning class-level prototypes and contrasting them across classes, PALM’s loss function encourages intra-class compactness and inter-class separability at  prototype level.
Similarly, PFS~\citep{wu2024pursuing} introduces a clustering loss that further promotes tight feature clustering within each ID class, enhancing the consistency of ID representations.

\textbf{A special case: Long-tail ID data.} 
ID data in real scenarios may present a long-tailed distribution due to the difficulty of collection and frequency of occurrence.
This imbalance will strongly affect the performance of OOD detection.
Many approaches are proposed to address the challenge of ID imbalance and enhance OOD detection capabilities.
OLTR~\citep{liu2019large}, POP~\citep{mehta2022out}, and AREO~\citep{sapkota2022adaptive} each address long-tailed recognition from different perspectives. OLTR enhances tail class robustness by leveraging visual memory to bridge head and tail embeddings. POP focuses on medium-to-tail classes through a prototype-based mixing strategy, generating mixed samples to improve class separation. In contrast, AREO models sample uncertainty via evidential learning and dynamically adjusts training using a multi-scheduler mechanism to better balance majority and minority classes.

Most existing OOD detection methods assume a uniform probability distribution between OOD and ID samples. To address class imbalance, \citet{jiang2023detecting} recalibrate OOD scores by incorporating class priors and KL divergence, improving robustness under skewed distributions. Open-Sampling~\citep{wei2022open} rebalances ID class priors using noisy OOD labels sampled from a complementary distribution. In contrast, COOD~\citep{hogeweg2024cood} ensembles multiple OOD measures via a supervised model, effectively mitigating individual method limitations and handling data imbalance.
\begin{figure}[t]
    \centering
    \includegraphics[width=0.8\textwidth]{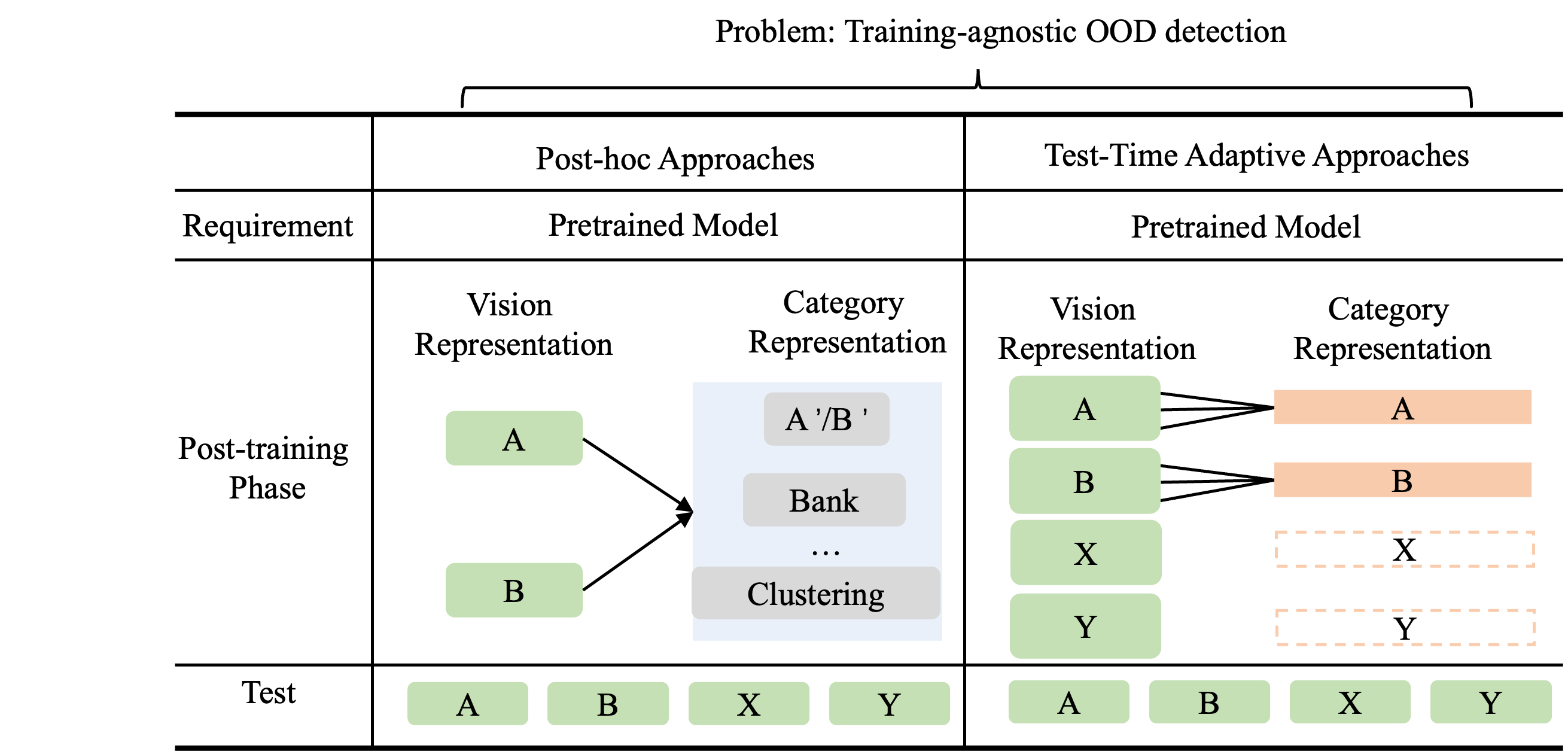}
    \caption{Illustration of training-agnostic OOD detection approaches. Both methods require access to a pre-trained model. Post-hoc approaches do not involve any operations during the post-training phase, while test-time adaptive approaches necessitate adaptation based on the samples encountered during testing. ``A'/B' '' means that the original features are deformed; ``Bank'' means that some samples are stored; ``Clustering'' means that clustering is performed for ID images.}
    \label{fig:training-agnostic}
\end{figure}
\subsection{OOD Detection Approaches with Both ID and OOD Data }\label{Approaches with Both ID and OOD Data}
\textbf{Overview.} 
In some known deployment scenarios, real OOD data can be easily collected at a low cost.
Some methods based on this assumption focus on how to use OOD data for better detection performance.
Differing from methods involving OOD Synthesis, in these research directions, models have access to real-world OOD data during the training phase. 
The primary focus in such problems is on optimizing the model's decision boundary, rather than the OOD data itself.
Due to the introduction of real OOD information, the decision boundary between ID and OOD data is further accurately calculated.

\textbf{Boundary Regularization.} The Boundary Regularization class of methods belongs to the traditional Outlier Exposure (OE) approaches. The central idea of ~\citet{hendrycks2018deep} and ~\citet{hein2019relu} are to fully leverage OOD data to optimize the model's decision boundary, thus achieving OOD detection.
Proponents of this concept can utilize auxiliary anomaly datasets to enhance the OOD detector, enabling it to generalize and detect anomalous information not encountered during training. 
The central idea of this method can be grasped from Fig. ~\ref{fig:Training} (e).


Specifically, given a model \( f \) and the original loss function, the model training process aims to minimize the objective:
\begin{equation}
    \mathbb{E}_{(x, y) \sim \mathcal{D}_{\text{in}}} \left[ \mathcal{L}(f(x), y) + \lambda \mathbb{E}_{x' \sim \mathcal{D}_{\text{out}}^{\text{OE}}} \left[ \mathcal{L}_{\text{OE}}(U, f(x)) \right] \right],
\end{equation}
over the parameters of \( f \), where $x'$ represents auxiliary anomaly data and $U$ denotes the uniform distribution. $\mathcal{L}_{\text{OE}}$ represents the cross-entropy loss with respect to $U$. Here, $\mathcal{D}_{\text{in}}$ denotes the distribution of ID data, while $\mathcal{D}_{\text{out}}^{\text{OE}}$ denotes the distribution of OOD.

The fundamental purpose is to compel the model to optimize the OOD data distribution to a uniform distribution, a principle that is universal in OE-type approaches. 
The specific design of $\mathcal{L}_{\text{OE}}$ can depend on other task requirements and the chosen OOD score.
This design can utilize a maximum softmax probability baseline ~\citep{hendrycks2016baseline} detector to detect anomalous data.
Compared to traditional softmax scores, EnergyOE ~\citep{liu2020energy} builds upon OE by leveraging energy scores for better discrimination between ID and OOD samples, and it is less prone to issues of overconfidence.
Specifically, its calculation formula: 
\begin{equation}
    E(x;f)=-T \cdot log\displaystyle\sum_{i}^{K}{e}^{{f}_{i}(x)/T},
\end{equation}
where the temperature coefficient \( T \) is used and \( f(x) \) denotes the discriminative neural classifier \( f(x): \mathbb{R}^D \rightarrow \mathbb{R}^K \), which maps an input \( x \in \mathbb{R}^D \) to \( K \) real-valued logits.

~\citet{mohseni2020self} train a model using a self-supervised approach, optimizing the objective function for unlabeled OOD samples using pseudo-labeling to generalize OOD detection capabilities.
~\citet{vyas2018out} similarly employ self-supervised training of the classifier. Unlike the OE approach, its aim is to find a gap between the average entropy of OOD and ID samples.
MixOE ~\citep{zhang2023mixture} takes into account the beneficial effect of subtle OOD samples on enhancing the generalization ability of OOD detection. Its main idea is to mix ID and OOD data samples to broaden the generalization of OOD data.
Training the model with these outliers can linearly decrease the prediction confidence with inputs from ID to OOD samples, explicitly optimizing the generalization ability of the decision maker.

\textbf{Outlier Mining.} The traditional OE concept assumes the existence of ID input $\mathcal{D}_{\text{in}}$ and OOD input $\mathcal{D}_{\text{out}}$, both independently and heterogeneously distributed, originating from different sources.
However, this premise cannot be fully guaranteed in the current training process due to potential noise in the training OOD data. 
Outlier Mining differs slightly from the traditional OE approach in that, although it also utilizes real-world OOD samples to address the issue, it focuses on identifying the optimal selection within the existing OOD data.
The main process is depicted in Fig. ~\ref{fig:Training} (d).
Different approaches have been proposed for selecting representative OOD samples: POEM ~\citep{ming2022poem} utilizes posterior sampling to identify high boundary score anomalies, while ~\citet{li2020background} employs data resampling with priority score reweighting for hard negative instances. In contrast, DAOL ~\citep{wang2023learning} models OOD distribution using Wasserstein balls to select challenging OOD samples based on the disparity between real and auxiliary data.

Beyond solely relying on raw data, another direction in addressing this issue involves synthesizing representative outlier data by utilizing authentic OOD data through information extrapolation.
DivOE ~\citep{zhu2023diversified} introduces a novel learning objective to alleviate challenges associated with limited auxiliary OOD datasets.
It achieves this by adaptively inferring and learning information from surrogate OOD data through the maximization of differences between generated OOD data and original data, given the specified anomalies.
This adaptive inference extends to a broader spectrum, addressing the limitations imposed by a finite auxiliary OOD dataset.
Moreover, DOE ~\citep{wang2023out} introduces a Min-Max learning strategy to identify the most challenging OOD data for a synthetic model.
Through model perturbation, the data is implicitly transformed, and the model continues learning from this perturbed data to improve its robustness.


\textbf{Long-tail ID data.} Recent studies have explored OOD detection in scenarios with imbalanced ID training data. Different approaches have been proposed to address this challenge: PASCAL ~\citep{wang2022partial} uses contrastive loss to separate tail-class from OOD data, while COCL ~\citep{miao2023out} introduces a learnable tail class prototype to better distinguish between them. ~\citet{choi2023balanced} focuses on balancing cross-class distribution of auxiliary OOD data through energy regularization, and EAT ~\citep{wei2023eat} expands the classification space by dynamically assigning virtual labels to OOD data during training.

Despite the success and considerable attention received by methods like Outlier Exposure in the research community, there are voices questioning the essence of allowing access to OOD data during training.
Nevertheless, concerns are raised that the superior classification performance observed in certain datasets may not necessarily translate to competitiveness in real-world deployment, challenging the original intention of OOD detection.

\section{Problem: Training-agnostic OOD Detection}\label{sec:Training-agnostic OOD Detection}
\textit{Training-agnostic OOD detection} focuses on adaptation strategies at test time for models that already possess downstream classification capabilities, as opposed to the focus on classifier performance seen in training-driven OOD detection.
Based on whether they rely on dependencies among test data, methods are categorized into two types: post-hoc and test-time adaptive approaches, as illustrated in Fig.~\ref{fig:training-agnostic}. 
Post-hoc methods compute results for individual samples independently, unaffected by changes in other samples. 
In contrast, test-time adaptive methods leverage information shared among test samples to enhance OOD detection.

\subsection{Post-hoc OOD Detection Approaches}\label{post-hoc}
\textbf{Overview.} 
Given a well-trained model, this problem scenario involves utilizing only the intermediate results computed by the trained model during testing, without modifying any parameters of the model, to accomplish the OOD detection task.
Post-hoc methods are favored for their lightweight nature, low computational costs, and the fact that they require minimal modifications to the model and objectives. 
Its main objective is to construct an effective scoring function that can accurately reflect the behavior of ID data.
These characteristics make them highly desirable for convenient and straightforward deployment in practical scenarios.
Post-hoc approaches are categorized into five types: \textit{Output-based}, \textit{Distance-based}, \textit{Gradient-based}, \textit{Feature-based} and \textit{Density-based}.
Recent work on this type of problem has some recent progress. A summary of the key factors involved in such methods is given in Table ~\ref{components}.

\begin{table}[t]
\caption{Comparison of key components in OOD detection methods.}
\label{components}
\small   %
\resizebox{0.7\linewidth}{!}{
\begin{tabular}{c|c|c|c|c|c}
\hline\hline
\multirow{2}{*}{\textbf{Type}} & \multirow{2}{*}{\textbf{Method}} & \multicolumn{4}{c}{\textbf{Space}} \\ \cline{3-6} 
 &  & \textbf{feature} & \textbf{logit} & \textbf{gradient} & \textbf{probability} \\ \hline
\multirow{4}{*}{Output-Based} & MSP~\citep{hendrycks2016baseline} &  &  &  & $\checkmark$ \\ 
 & Maxlogits~\citep{hendrycks2019scaling} &  & $\checkmark$ &  &  \\ 
 & Energy~\citep{liu2020energy} &  & $\checkmark$ &  &  \\  
 & GEN~\citep{liu2023gen} & $\checkmark$ &  &  & $\checkmark$ \\ \hline
\multirow{4}{*}{Distance-Based} & Mahalanobis~\citep{lee2018simple} & $\checkmark$ &  &  &  \\ 
 & NNGuide~\citep{park2023nearest} & $\checkmark$ &  &  &  \\ 
 & KNN~\citep{sun2022out} & $\checkmark$ &  &  &  \\ 
 & SSD~\citep{sehwag2021ssd} & $\checkmark$ &  &  &  \\ \hline
\multirow{5}{*}{Gradient-Based} & Grad~\citep{lee2020gradients} &  &  & $\checkmark$ &  \\ 
 & GradNorm~\citep{huang2021importance} & $\checkmark$ &  & $\checkmark$ & $\checkmark$ \\ 
 & GradOrth~\citep{behpour2023gradorth} & $\checkmark$ &  & $\checkmark$ &  \\ 
 & GAIA~\citep{chen2023gaia} & $\checkmark$ &  & $\checkmark$ &  \\ 
 & OPNP~\citep{chen2024optimal} & $\checkmark$ &  & $\checkmark$ &  \\ \hline
\multirow{11}{*}{Feature-Based} & ODIN~\citep{liang2017enhancing} & $\checkmark$ &  & $\checkmark$ & $\checkmark$ \\ 
 & ReAct~\citep{sun2021react} & $\checkmark$ &  &  &  \\ 
 & VRA~\citep{xu2023vra} & $\checkmark$ &  &  &  \\
 & Vim~\citep{wang2022vim} & $\checkmark$ & $\checkmark$ &  &  \\
 & Neco~\citep{ammar2023neco} & $\checkmark$ & $\checkmark$ &  &  \\ 
 & ASH~\citep{djurisic2022extremely} & $\checkmark$ &  &  &  \\ 
 & Optimal-FS~\citep{zhao2024towards} & $\checkmark$ & $\checkmark$ &  &  \\ 
 & SCALE~\citep{xu2023scaling} & $\checkmark$ &  &  &  \\ \hline
\multirow{2}{*}{Density-Based} & ConjNorm~\citep{peng2024conjnorm} & $\checkmark$ & $\checkmark$ &  &  \\ 
 & GEM~\citep{morteza2022provable} & $\checkmark$ & $\checkmark$ &  & $\checkmark$ \\ \hline\hline
\end{tabular}
}
\end{table}

\textbf{Output-based.} Algorithms based on output primarily aim to explore the latent representations of the output from the intermediate layers of neural networks, which include logits and class distributions, among others.
MSP ~\citep{hendrycks2016baseline} is the first to employ the maximum softmax value to validate OOD detection effectiveness. For OOD samples, their output probability distribution tends to be closer to a uniform distribution, demonstrating the model's inability to correctly classify the category.
Departing from the MSP method, MaxLogits~\citep{hendrycks2019scaling} detects OOD samples using the maximum logit value, while Energy~\citep{liu2020energy} employs an energy-based score related to input probability density. Compared to MaxLogits, Energy is less sensitive to model overconfidence and provides a more theoretically grounded metric.

GEN ~\citep{liu2023gen} introduces the concept of generalized entropy and directly utilizes Bregman divergence to compute the statistical distance between the model's probability output and uniform distribution, aiming to identify OOD data.
In addition, leveraging sufficient prior knowledge might be a viable solution.
ZODE ~\citep{xue2024enhancing} performs predictions on samples across multiple pre-trained models simultaneously to determine whether multiple models can identify OOD samples, using this as a basis to distinguish between data.
LogicOOD ~\citep{kirchheim2024out} presents a novel approach that uses first-order logic for knowledge representation to perform OOD detection.
This reasoning system uses prior knowledge to infer whether an input is consistent with prior knowledge about the training distribution.
It is particularly user-friendly in terms of interpretability, as it allows comparing the output of samples against a knowledge base to determine whether they belong to the ID data.

\textbf{Distance-based.} Another approach in OOD detection research focuses on measuring statistical distance metrics.
Mahalanobis ~\citep{lee2018simple} is typically computed by calculating the distance between the feature vector and its mean.
Specifically, for each class, we compute the mean and covariance matrix of its feature vectors. During testing, it calculates the Mahalanobis distance between the feature vector and the mean of each class.
Mahalanobis++~\citep{mueller2025mahalanobis++} also uses Mahalanobis distance, but applies L2 norm normalization to the features in the early stages of calculation to eliminate the instability in feature extraction across different models, which affects the distribution assumption they rely on.
SSD ~\citep{sehwag2021ssd} essentially utilizes the Mahalanobis distance.
After being trained on unlabeled ID data via self-supervised representation learning, it employs the Mahalanobis distance as a statistical measure for classification using the pre-trained model.
In comparison, while the Mahalanobis distance makes strong distributional assumptions about the data, KNN ~\citep{sun2022out} explores the effectiveness of non-parametric nearest-neighbor distance for OOD detection.
By measuring the k-nearest neighbor distance between input embeddings and training set embeddings, a threshold is designed to determine whether the data belongs to the ID data.
NNGuide ~\citep{park2023nearest} takes a step further in the direction of granularity by combining the idea of KNN.
It assigns weights before the traditional OOD Score, depending on the nearest neighbor distance between the sample and the embeddings in the training set.

\textbf{Gradient-based.} Gradient-based methods leverage model gradients to quantify uncertainty. Grad~\citep{lee2020gradients} and GradNorm~\citep{huang2021importance} both utilize gradient magnitudes, with the latter specifically comparing KL-divergence-based gradients between ID and OOD samples. GradOrth~\citep{behpour2023gradorth}, instead of focusing on overall gradient norms, projects gradients onto low-rank subspaces to capture OOD features. GAIA~\citep{chen2023gaia} introduces channel-wise and zero-deflation abnormality scores to assess distribution shifts without prior knowledge. OPNP~\citep{chen2024optimal} further improves OOD detection by pruning parameters and neurons near zero, thus enhancing generalization.
In recent research, PRO~\citep{chen2025leveraging} introduces an adversarial score, with the core idea being to perturb the input data using gradient descent to search for the local minimum score near the original input. This method effectively lowers the confidence of OOD samples and is compatible with existing MSP detection methods during score calculation, thereby enhancing detection performance.
S\&I~\citep{zhang2025splitting} identifies the issue of insufficient sample gradient feature discrimination in this type of method. By introducing adversarial samples and integrating gradients along the attribution path, it achieves a more accurate feature distinction strategy.

\textbf{Feature-based.} Research in this direction explores the role of intermediate neural features for OOD detection. ODIN~\citep{liang2017enhancing}, inspired by adversarial examples, demonstrates that small input perturbations can enhance OOD detection performance.
The perturbation formula is as follows:
\begin{equation}
    \tilde{x} = x + \varepsilon  sign(\bigtriangledown_{x}logmax_{c}p_{c}(x)),
\end{equation}
where the parameter $\varepsilon$ is the perturbation magnitude.
For a given input $x$, compute its logit output $p_{c}(x)$.
Therefore, ReAct ~\citep{sun2021react} focuses on the high activation values in the intermediate results of the model. 
These activation values do not affect the model's classification, but truncating high activation values can significantly improve OOD detection performance.
VRA ~\citep{xu2023vra}, an extended iteration of ReAct 
 ~\citep{sun2021react}, builds upon the premise of ReAct which truncates only high activation values. 
However, VRA posits that this might not be the optimal solution and, therefore, employs a variational approach to seek an optimal solution.
It utilizes piecewise functions to emulate suppression or amplification operations, aiding the model in recognizing anomalous data.
KANs~\citep{canevaro2025advancing} also focus on the neuron activation states. Due to the local neuroplasticity characteristic of Kolmogorov-Arnold Networks~\citep{liu2024kan}, KANs distinguish between ID and OOD data by comparing the activation response differences between the trained and untrained KANs.
SHE~\citep{zhang2022out} introduces a storage-based approach by leveraging Hopfield energy~\citep{krotov2016dense} to facilitate OOD detection, averaging logits per class to form a reference pattern. Building on channel selection, DDCS~\citep{yuan2024discriminability} evaluates and corrects neural network channels according to their discriminative ability, using inter-class similarity and variance. Similarly, LINe~\citep{yong2023line} focuses on neuron-level feature outputs, employing Shapley value pruning to retain only the most informative neurons for OOD detection.
CADRef~\citep{ling2025cadref} decouples the features by using a strategy based on the sign alignment between the relative features and model weights, dividing the sample features into positive and negative errors. Intuitively, the positive and negative error components influence the final output logits, which helps in identifying OOD images from this perspective.
ITP~\citep{xu2025itp} optimizes the model by quantifying the contribution of model parameters to ID data prediction, specifically by evaluating the contribution of each parameter through its partial derivatives and removing those with low contribution. Additionally, during the testing phase, ITP uses a right-tail Z-score test to assess whether any parameter exhibits overconfidence in the classification of a given test sample, further refining the model's decision-making.


Feature shaping refines intermediate features during forward propagation, offering a simple and effective way to enhance OOD detection without affecting original classification results. Building on this, ViM~\citep{wang2022vim} integrates features, logits, and probabilities by constructing virtual logits, and leverages the penultimate layer’s null space—irrelevant to classification yet highly effective for OOD detection.
Its computation of score $S$ can be expressed in the following form:
\begin{equation}
    S = -\alpha \|z^{P^\perp}\|^2 + \text{LogSumExp} f(z),
\end{equation}
where $\alpha$ is a scaling constant, computed by the model.
Here $z = z^P + z^{P^\perp}$ and $z^{P^\perp}$ is the projection of $z$ to $P^\perp$.
And it have $Wz^{P^\perp} = 0$.
Additionally, $LogSumExp$ represents the computation process of the energy function ~\citep{liu2020energy}, and $f(z)$ represents the logit output of the model.
The first term here represents virtual logits, while the second term represents the score of the energy function.
Neco ~\citep{ammar2023neco} subsequently reveals the prevalent phenomenon of neural collapse in contemporary neural networks, impacting OOD detection performance.
The observation of orthogonal trends between ID data and OOD data features is leveraged to differentiate OOD data.
Similarly, NCI~\citep{liu2023detecting} studies the phenomenon of neural collapse and discovers that during training, ID samples tend to cluster around class weight vectors in the penultimate feature space, while OOD samples do not exhibit this clustering behavior. Therefore, NCI leverages the proximity between features and class weight vectors to effectively distinguish ID samples from OOD samples, while maintaining low computational cost.
ASH~\citep{djurisic2022extremely} and neuron activation pruning focus on direct modification or removal of activations, with ASH offering a simple dynamic scheme and pruning methods building upon its foundation. In contrast, NAC~\citep{liu2023neuron} introduces neuron activation coverage as a statistical indicator, aiming to capture OOD likelihood based on rarely activated neurons. BLOOD~\citep{jelenic2023out} and related methods leverage the smoother intermediate representations of ID data, using this property to design new statistical measures for OOD discrimination. Feature shaping approaches~\citep{zhao2024towards} partition the feature space and estimate piecewise approximations to logits, while SCALE~\citep{xu2023scaling} highlights the importance of scaling metrics and observes lower pruning rates for OOD samples. Overall, these methods differ in whether they reshape activations directly, exploit statistical properties, or design new metrics for improved OOD detection.

\textbf{Density-based.} Recent density-based OOD detection methods have achieved notable improvements by more effectively modeling the true data distribution.
For example, GEM~\citep{morteza2022provable} employs class-conditional Gaussian distributions with statistical metrics for model validation. Building upon this, ConjNorm~\citep{peng2024conjnorm} extends the framework to exponential family distributions using Bregman divergence, enabling broader applications.



\subsection{Test-Time Adaptive OOD Detection Approaches}\label{test-time}
\textbf{Overview.} 
Recently, many works~\citep{yu2024stamp,yu2023benchmarking,sun2020test,yu2025test,liang2025comprehensive} have focused on test-time adaptive methods, where a model, trained on the training set, adapts during the testing phase.
In the context of OOD detection, these methods aim to leverage unlabeled test data—whether the complete test set or mini-batches—to improve performance through model adaptation.
Test-time adaptive approaches are based on the theoretical insight ~\citep{fang2022out} that detecting OOD samples using only ID samples without any additional knowledge 
can be challenging and potentially limited.
These methods can be divided into two categories according to whether the model be modified during the testing time: \textit{Model-optimization-based} and \textit{Model-optimization-free}.
Both of them undergo a post-training phase, during which the trained model can be adapted, regardless of whether it is updated.

\begin{figure}[t]
    \centering 
    \includegraphics[width=0.8\linewidth, keepaspectratio]{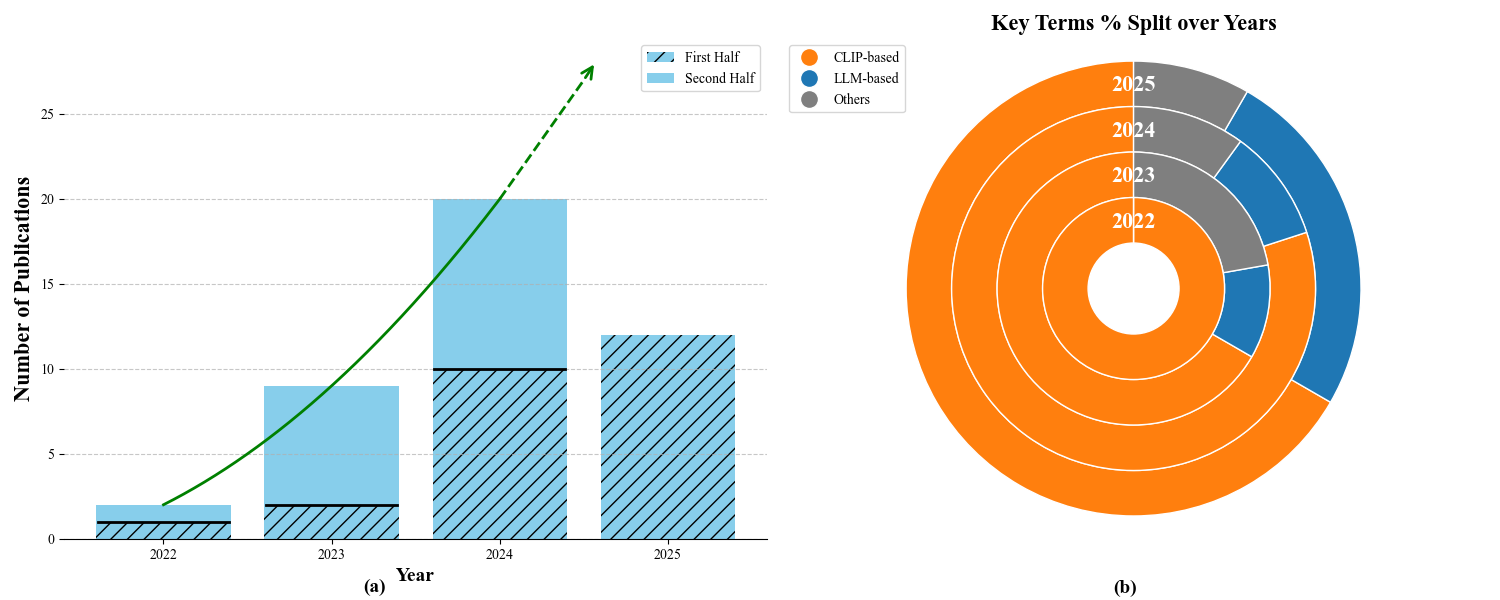}
    \caption{Overview and trends in LPM-based OOD Detection Methods.
    (a) presents statistical data on influential research works in recent years, which involves papers indexed in prominent journals or conferences in the field. In the bar chart, the lower and upper parts represent the total number of papers in the first and second halves of each year, respectively. (b) shows a trend pie chart illustrating the proportion of keyword categories related to the research works in that year.}
    \label{fig:trends}
\end{figure}

\textbf{Model-optimization-based.} 
Model optimization-based methods enhance the trained model by leveraging unlabeled data during the post-training phase.

A line of methods~\citep{woods, du2024does} leverage “wild data”—a mix of unlabeled ID and OOD samples—to enhance OOD detection. WOODS~\citep{woods} focuses on cleaning wild data for reliable OOD regularization, while SAL~\citep{du2024does} analyzes its impact through separability and learnability. However, wild data may introduce unintended information if it differs from test OOD, and additional training requirements can increase computational costs.

Another line of model-optimization-based methods draw inspiration from semi-supervised-learning techniques ~\citep{zhu2005semi} for efficient post-training. While basic pseudo-labeling ~\citep{arazo2020pseudo} labels test data directly, AUTO ~\citep{yang2023auto} uniquely focuses on pseudo-OOD data with semantic consistency. In contrast, ATTA ~\citep{gao2023atta} and SODA ~\citep{geng2023soda} utilize both pseudo-ID and pseudo-OOD data, with SODA implementing dual-loss optimization and ATTA applying differential weighting strategies.

\textbf{Model-optimization-free.} 
Modifying the original trained model is infeasible in certain security-sensitive scenarios. 
Therefore, methods enabling test-time adaptation without requiring model updates, termed ``model-optimization-free'' techniques, are increasingly garnering interest. 
These approaches enhance the utilization of test data by either memorizing it or incorporating additional modules on top of the original model.
Both ETLT ~\citep{test-time_ETLT} and GOODAT ~\citep{wang2024goodat} preserve model integrity by training add-on modules for OOD score adjustment. ETLT leverages the linear correlation between features and OOD scores, implementing both offline and online variants, while GOODAT develops a graph-specific masker with GIB-boosted losses. In contrast, AdaOOD ~\citep{AdaOOD} and OODD ~\citep{yang2025oodd} adopt non-parametric approaches, utilizing memory banks with k-nearest neighbours and dynamic dictionaries, respectively.

\textbf{Online v.s. Offline.} 
Most post-hoc methods ~\citep{liang2018enhancing, test-time_ETLT} emphasize the offline scenario, where OOD detectors remain static and fixed after deployment. 
In contrast, the majority of test-time methods~\citep{AdaOOD, gao2023atta} adopt the online scenario to obtain the decision boundary dynamically, minimizing the risk of incorrect OOD predictions at each time step.

\textbf{More challenging scenario.} 
In the context of test-time OOD detection scenarios, some researchers have proposed more challenging configurations that demand a higher level of capability from the models.
MOL~\citep{wu2023metaMOL} introduces a more realistic problem scenario, namely Continuous Adaptive Out-of-Distribution (CAOOD) detection, 
aimed at addressing the challenge of constantly changing ID and OOD distributions in the real world.  
The meta-learning approach is employed to swiftly adapt models in response to the complexities encountered in various scenarios in MOL.


\section{Problem: Large Pre-trained Model-based OOD Detection}\label{sec:Large Pre-trained Model based OOD Detection}

\begin{figure}[t]
    \centering
    \includegraphics[width=0.85\textwidth]{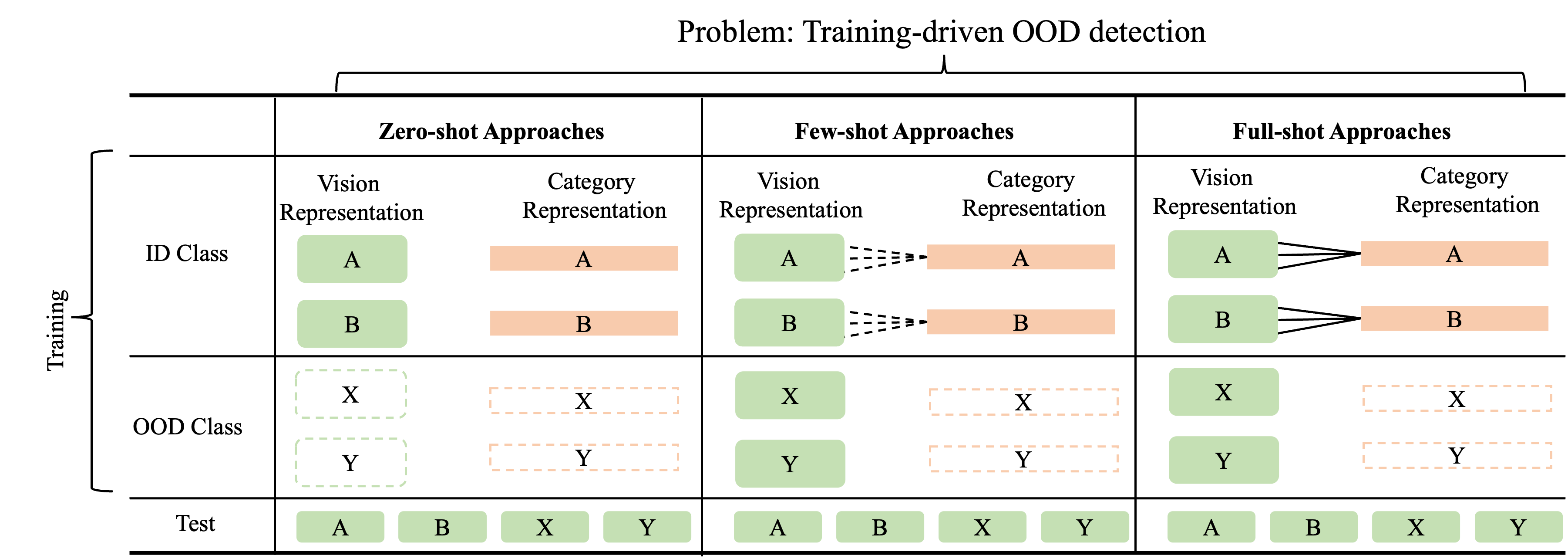}
    \caption{Illustration of large pre-trained model-based OOD detection approaches. In the training phase, zero-shot approaches require only category labels of ID classes. Few-shot approaches need a subset of images of each ID class along with the category labels (indicated by dashed lines). Full-shot approaches utilize both the category labels and all images of each ID class. None of these approaches use labels or images of OOD categories.}
    \label{fig:LPM}
\end{figure}

Large pre-trained models have showcased remarkable performance in numerous downstream ID classification tasks, 
but their potential in OOD detection tasks remains a less-explored area.
Recent research ~\citep{vaze2021open} highlights a correlation between higher ID classification accuracy and better OOD detection performance.
Consequently, large pre-trained model-based (LPM-based) OOD detection problem comes naturally.
In recent years, large pre-trained models of various types, including single-modal models (ViT~\citep{dosovitskiy2020image}, BERT~\citep{devlin2018bert}, Diffusion~\citep{ho2020denoising}), visual language models (VLMs) (CLIP~\citep{radford2021learning}, multi-modal Diffusion~\citep{saharia2022photorealistic}, ALIGN~\citep{jia2021scaling}, 
), and large language models (LLMs) (GPT3~\citep{brown2020language}), have been increasingly utilized for OOD detection tasks, as shown in Fig.~\ref{fig:trends}. Based on the trend chart, it can be anticipated that LPM-based OOD Detection will continue to be a key area of focus in the future.
Leveraging the powerful representational capabilities of large pre-trained models has further relaxed the constraints of OOD detection tasks, leading to a focus on more challenging and realistic scenarios, which has emerged as a new hotspot.
Given the number of ID shots exposed to the large pre-trained model, OOD detection in this section can be classified into 
Zero-shot, Few-shot, and Full-shot OOD detection, as shown in Fig.~\ref{fig:LPM}.
The performance evaluations of several relevant competitive methods are summarized in Table ~\ref{tab:comparison} to provide an understanding of the performance level of OOD detection in this area.



\vspace{-5pt}
\subsection{Zero-shot OOD Detection Approaches}\label{zero-shot-ood}
\textbf{Overview.} 
Given the large pre-trained model and ID class names, we undertake the same task as the OOD detection, precisely detecting OOD data to abstain from prediction and accurately classifying ID data.
Note that we rely solely on textual category knowledge, without the need for access to ID images.
It's important to clarify that ``ID'' here refers to the ID of the specific downstream task, not the dataset ID data during pre-training. 
Existing research on zero-shot OOD detection using vision-language models (\textbf{VLMs-based}) can be broadly categorized into two main approaches, depending on the underlying models employed: diffusion-based and CLIP-based methods.

\textbf{Diffusion-based.}
Recent generative approaches, such as diffusion models~\citep{ho2020denoising}, effectively capture the ID data by modeling its underlying distribution. These models typically detect OOD samples by evaluating the likelihood of a test input belonging to the modeled ID distribution. For instance, RONIN~\citep{nguyen2024zero} leverages a diffusion model for ID inpainting, combined with the CLIP model to compute similarity scores for OOD detection. However, most existing works focus more extensively on CLIP-based methods due to their superior zero-shot capabilities and flexibility.

\textbf{CLIP-based.}
In contrast to diffusion-based approaches, CLIP-based methods  have received significant attention for zero-shot OOD detection. These methods exploit the powerful vision-language alignment capability of CLIP to perform OOD detection without requiring additional training on OOD data. In the following, we focus on the development and evolution of CLIP-based zero-shot OOD detection methods, discussing both approaches that require extra training and post-hoc methods, as well as recent advances leveraging textual information.

A representative approach, ZOC~\citep{esmaeilpour2022zeroZOC}, introduces an image description generator trained on large-scale image captioning datasets~\citep{lin2014microsoft}, enabling the generation of candidate unseen labels for OOD detection. Notably, ZOC treats CLIP primarily as a feature extractor and does not inherently endow it with OOD discrimination capabilities. Building upon this, CLIPN~\citep{wang2023clipn} enhances CLIP’s expressiveness by incorporating a ``no-prompt’’ encoder, thereby empowering CLIP to explicitly reject unfamiliar inputs—a critical step toward robust OOD detection in open-world scenarios. Beyond methods that require further training, a parallel line of research investigates post-hoc zero-shot OOD detection, aiming to maximize CLIP’s potential without additional fine-tuning. The most straightforward baseline in this category computes the normalized text-image similarity as the OOD score. To improve upon this, MCM~\citep{ming2022delvingMCM} replaces the similarity score with a Maximum Concept Matching (MCM) score, providing a theoretically grounded and empirically robust alternative. The MCM score is defined as:
\begin{equation}
S_{MCM}(x'; Y_{in}, T, \tau) = \max_i \frac{e^{s_i(x')/\tau}}{\sum_{j=1}^{K} e^{s_j(x')/\tau}}.
\end{equation}
where  $x’$  denotes the test image,  $Y_{in}$  is the set of ID labels,  $T(t_i)$  is the text embedding for prompt  $t_i$ ,  $s_i(x’)$  is the cosine similarity between the image and text features, and $\tau$  is the softmax temperature.

Several works have enriched the textual context in CLIP-based OOD detection by augmenting candidate labels with curated corpora, as seen in NegLabel~\citep{anonymous2024negative}, LAPT~\citep{zhang2024lapt}, CSP~\citep{chen2024conjugated}, and CLIPScope~\citep{fu2024clipscope}.
Specifically, NegLabel selects negative labels from a corpus that are semantically dissimilar to ID classes, thereby sharpening the model’s ability to distinguish OOD samples. CLIPScope~\citep{fu2024clipscope} goes a step further by integrating these negative labels into a Bayesian scoring framework to revise the original OOD score. LAPT, in contrast, utilizes text-to-image generation or image retrieval models to obtain visual exemplars for both ID and negative labels, followed by prompt tuning to optimize OOD discrimination. It is worth emphasizing that methods such as ZOC~\citep{esmaeilpour2022zeroZOC}, CLIPN~\citep{wang2023clipn}, NegLabel~\citep{anonymous2024negative}, and CLIPScope~\citep{fu2024clipscope} each propose distinct OOD scoring mechanisms tailored to their respective architectures. In contrast, the MCM score can be seamlessly integrated into diverse CLIP-based frameworks.

Recent zero-shot OOD detection methods mainly differ in negative sampling, noise handling, and semantic grouping. AdaNeg~\citep{zhang2024adaneg} and AdaND~\citep{cao2025noisy} refine negative sampling and noise separation, while SimLabel~\citep{zou2025simlabel} focuses on automatic semantic grouping. For outlier representation, OLE~\citep{ding2024zero} expands outlier label diversity through clustering and interpolation, and prompt learning~\citep{jung2024enhancing} introduces new scoring functions for post-hoc detection without retraining. Matrix-based methods further diversify the field, with SeTAR~\citep{li2024setar} applying low-rank approximations and OT-DETECTOR~\citep{liu2025ot} using optimal transport to capture semantic gaps.


\textbf{LLM-based.} 
With the rapid advancement of LLMs, new opportunities have emerged in the field of OOD detection by leveraging the extensive world knowledge and powerful semantic reasoning capabilities of LLMs. The primary strength of LLMs lies in their ability to provide comprehensive and context-rich descriptions of ID labels, which can be crucial for distinguishing between ID and OOD samples. However, a significant challenge in employing LLMs for OOD detection is their tendency to generate hallucinations—false or misleading information—which can undermine the reliability of OOD identification.

Recent works explore the relationship between visual and semantic spaces to improve OOD detection. Studies~\citep{huang2024out, dai2023exploringLLM} leverage the distinct properties of visually similar categories in semantic space. \citet{dai2023exploringLLM} develops a consistency-based method with object detection to enhance LLM-based class descriptions, while \citet{huang2024out} proposes ODPC to generate peer classes using both ID labels and OOD samples. These approaches differ in their scoring mechanisms, with \citet{dai2023exploringLLM} using MSP and ODPC employing KNN-based scoring.
Building upon these foundations, subsequent studies have systematically investigated the behavior of LLMs in OOD detection. For example, \citet{liu2023good} analyze the propensity of LLMs to detect near-OOD versus far-OOD samples, the impact of different fine-tuning strategies, and the suitability of various OOD scoring functions. 

More recently, LLM-based approaches such as CMA~\citep{lee2025concept}, COOD~\citep{liu2024cood}, and EOE~\citep{cao2024envisioning} have extended OOD detection into multi-label and concept-rich regimes. These methods utilize LLMs not only as knowledge generators, but also as proxy creators and synthetic data generators~\citep{abbas2025out}, thereby enriching the semantic context available for OOD detection. The synergy between VLMs for visual grounding and LLMs for semantic reasoning has been further highlighted by works such as ReGuide~\citep{kim2024reflexive}, which propose self-guided, image-adaptive concept generation to enhance robustness. Collectively, these advancements demonstrate the growing potential of LLM-based techniques to address the complex challenges of OOD detection by effectively integrating visual and semantic information.

\textbf{Remark.} 
As mentioned above, ``zero-shot'' here refers to no exposure to ID images and only access to ID labels.
While some works~\citep{fort2021exploring,ding2024zero} improve performance by using known OOD class names as candidate labels, this approach relies on the often unrealistic assumption that OOD labels are readily available.
This setting is also distinct from recent unsupervised fine-tuning approaches~\citep{liang2023realistic}, which address a more realistic scenario where the unlabeled data for adaptation is possibly contaminated with OOD samples. 




\begin{table}[]
\caption{Performance evaluation of some competitive methods based on CLIP-ViT-B/16 using the ImageNet-1K dataset as the ID dataset and iNaturalist, SUN, Places, and Textures as OOD datasets.  The results are cited from~\cite{ming2022delvingMCM,chen2024conjugated,zhang2024adaneg,li2024learning} and the best result is emphasized in bold. }
\label{tab:comparison}
\resizebox{\textwidth}{!}{
\begin{tabular}{ll|ll|ll|ll|ll|ll}
\hline
\hline
            \multirow{2}{*}{Scenario}    &     \multirow{2}{*}{Method}       & \multicolumn{2}{c|}{iNaturalist}   & \multicolumn{2}{c|}{SUN}           & \multicolumn{2}{c|}{Places}        & \multicolumn{2}{c|}{Textures}      & \multicolumn{2}{c}{Average}       \\ \cline{3-12} 
                           &           & \multicolumn{1}{l}{AUROC $\uparrow$ } & FPR95 $\downarrow$ & \multicolumn{1}{l}{AUROC  $\uparrow$ } & PFR95 $\downarrow$ & \multicolumn{1}{l}{AUROC $\uparrow$} & FPR95 $\downarrow$ & \multicolumn{1}{l}{AUROC $\uparrow$} & FPR95 $\downarrow$ & \multicolumn{1}{l}{AUROC $\uparrow$} & FPR95 $\downarrow$ \\ \hline
\multirow{4}{*}{Zero-shot} & ZOC ~\citep{esmaeilpour2022zeroZOC}       & \multicolumn{1}{l}{86.09} & 87.30  & \multicolumn{1}{l}{81.20}  & 81.51 & \multicolumn{1}{l}{83.39} & 73.06 & \multicolumn{1}{l}{76.46} & 98.90  & \multicolumn{1}{l}{81.79} & 85.19 \\ 
                           & MCM ~\citep{ming2022delvingMCM}      & \multicolumn{1}{l}{94.59} & 32.20  & \multicolumn{1}{l}{92.25} & 38.80  & \multicolumn{1}{l}{90.31} & 46.20  & \multicolumn{1}{l}{86.12} & 58.50  & \multicolumn{1}{l}{90.82} & 43.93 \\ 
                           & CLIPN   ~\citep{wang2023clipn}   & \multicolumn{1}{l}{95.27} & 23.94 & \multicolumn{1}{l}{93.93} & 26.17 & \multicolumn{1}{l}{92.28} & 33.45 & \multicolumn{1}{l}{90.93} & 40.83 & \multicolumn{1}{l}{93.10}  & 31.10  \\ 
                           & NegLabel ~\citep{anonymous2024negative} & \multicolumn{1}{l}{99.49} & 1.91  & \multicolumn{1}{l}{95.49} & 20.53 & \multicolumn{1}{l}{91.64} & 35.59 & \multicolumn{1}{l}{90.22} & 43.56 & \multicolumn{1}{l}{94.21} & 25.40  \\ 
                           & LAPT ~\citep{zhang2024lapt} & \multicolumn{1}{l}{99.63} & 1.16  & \multicolumn{1}{l}{96.01} & 19.12 & \multicolumn{1}{l}{92.01} & 33.01 & \multicolumn{1}{l}{91.06} & 40.32 & \multicolumn{1}{l}{94.68} & 23.40  \\ 
                            & CSP ~\citep{chen2024conjugated} & \multicolumn{1}{l}{99.60} & 1.54  & \multicolumn{1}{l}{96.66} & 13.66 & \multicolumn{1}{l}{92.90} & 29.32 & \multicolumn{1}{l}{93.86} & \textbf{25.52} & \multicolumn{1}{l}{95.76} & \textbf{17.51}  \\ 
                            & AdaNeg ~\citep{zhang2024adaneg} & \multicolumn{1}{l}{\textbf{99.71}} & \textbf{0.59}  & \multicolumn{1}{l}{\textbf{97.44}} & \textbf{9.50} & \multicolumn{1}{l}{\textbf{94.55}} & 34.34 & \multicolumn{1}{l}{\textbf{94.93}} & 31.27 & \multicolumn{1}{l}{\textbf{96.66}} & 18.92 \\ 
                           
                           \hline
                           
\multirow{3}{*}{Few-shot}  & CoOp ~\citep{zhou2022learning}      & \multicolumn{1}{l}{93.77} & 29.81 & \multicolumn{1}{l}{93.29} & 40.83 & \multicolumn{1}{l}{90.58} & 40.11 & \multicolumn{1}{l}{89.47} & 45.00    & \multicolumn{1}{l}{91.78} & 51.68 \\ 
                           & LoCoOp ~\citep{miyai2023locoop}    & \multicolumn{1}{l}{96.86} & 16.05 & \multicolumn{1}{l}{95.07} & 23.44 & \multicolumn{1}{l}{91.98} & 32.87 & \multicolumn{1}{l}{90.19} & 42.28 & \multicolumn{1}{l}{93.52} & 28.66 \\ 
                           & NegPrompt ~\citep{li2024learning} & \multicolumn{1}{l}{98.73} & 6.32  & \multicolumn{1}{l}{95.55} & 22.89 & \multicolumn{1}{l}{93.34} & \textbf{27.60}  & \multicolumn{1}{l}{91.60}  & 35.21 & \multicolumn{1}{l}{94.81} & 23.01 \\ \hline \hline
\end{tabular}
}
\end{table}

\subsection{Few-shot OOD Detection Approaches}\label{few-shot-ood}
\textbf{Overview.} 
Given the large pre-trained model and a few ID data, we can adapt the model using the ID data and subsequently detect OOD test data.
Zero-shot OOD detection does not necessitate any training images, making it suitable for scenarios with high-security requirements. 
However, it may face challenges related to domain gaps with ID downstream data, which can limit the performance of zero-shot methods. 
Therefore, there are many few-shot methods employed in OOD detection, and their effectiveness is often superior to that of zero-shot OOD detection.

\textbf{Studies.} 
Fine-tuning large pre-trained models with limited ID samples is a common adaptation strategy. Recent studies~\citep{ming2024does, dong2023towardsDSFG} systematically evaluate fine-tuning and parameter-efficient fine-tuning (PEFT) methods for OOD detection in VLMs, particularly CLIP. Both works highlight the superior OOD detection performance of PEFT over traditional fine-tuning. The MCM score~\citep{ming2022delvingMCM} and prompt-based approaches are also shown to be effective. FLYP~\citep{kim2024comparison}, which mimics CLIP-style pretraining, further outperforms PEFT in zero-shot OOD detection. Notably, few-shot OOD detection using outlier examples~\citep{fort2021exploring} is not included here, as it does not align with the focus on ID samples.

\textbf{Few-shot OOD Detection.}
Few-shot OOD detection aims to leverage a limited amount of ID data to adapt LPMs, primarily through prompt learning or parameter-efficient fine-tuning. 
Early approaches such as CoOp~\citep{zhou2022conditional} and ZegCLIP~\citep{zhou2023zegclip} focus on optimizing the contextual words in prompts while keeping the backbone fixed, thus preserving the generalization ability of the pre-trained models. Building upon this, LoCoOp~\citep{miyai2023locoop} and IDPL~\citep{bai2023id} further enhance context vector learning from a near-ID perspective: LoCoOp maximizes entropy to separate ID-irrelevant local features (e.g., backgrounds) from textual embeddings of ID classes, while IDPL synthesizes ID-like outliers near the ID boundary and employs a diversity loss to encourage variance among sampled OOD candidates.
In addition, MMFT~\citep{kim2025enhanced} projects image and text features into a shared hyperspherical space and introduces a cross-modal alignment loss function to promote the alignment of image and text representations in the hyperspherical space.
However, these prompt-only methods may underutilize the rich information contained in image features. Recent advances such as GalLoP~\citep{lafon2024gallop} and Local-Prompt~\citep{zeng2025local} address this limitation by jointly optimizing global and local prompts.

Beyond prompt learning, recent methods address the limitations of conventional fine-tuning, such as the loss of OOD awareness, by merging original and adapted features~\citep{dong2023towardsDSFG} or introducing self-calibrated tuning frameworks that balance ID and OOD objectives~\citep{yu2024self}. Other approaches improve prompt optimization through gradient-based techniques~\citep{tong2025enhancing}. To mitigate overfitting under limited supervision, strategies such as plug-and-play adapters~\citep{chen2024dual}, multi-modal alignment~\citep{wang2025mitigating}, and boundary regularization with synthesized OOD features~\citep{sun2024clip} have been proposed. Further refinements include optimizing negative prompts to better distinguish ID/OOD boundaries~\citep{li2024learning}. Collectively, these methods enhance few-shot OOD detection by improving adaptation, alignment, and boundary modeling.

\textbf{Meta-learning based.} 
Meta-learning aims to devise a learning approach that enables rapid adaptation to new challenges ~\citep{hospedales2021meta}.
OOD-MAML ~\citep{jeong2020ood} adapt model-agnostic meta-learning (MAML) for few-shot OOD detection.
It generates OOD samples and incorporates them along with ID data for the adapted N-way K-shot task, which is divided into N sub-tasks, each focusing on K-shot OOD detection. 
The decision on whether test data is OOD is based on the outcomes of these fast and simple N sub-tasks.
In contrast, HyperMix ~\citep{mehta2024hypermix} advocates for employing a hypernetwork-based method to enhance sample augmentation without the necessity for extra outliers.
This is because classes not included in a specific meta-training task can act as OOD samples.
\vspace{-5pt}
\subsection{Full-shot OOD Detection Approaches}\label{full-shot-ood}
\textbf{Overview.} 
%
While this setup is generally less realistic than the first two (zero-shot and few-shot), we list them separately to ensure a comprehensive review of existing methods across the spectrum.
Given the full set of ID data and corresponding labels, VLMs can enhance OOD detection significantly by fine-tuning. 
Moreover, a novel task called ``PT-OOD'' detection is introduced.

\textbf{Fine-tuning based.} 
With access to the complete dataset, then more data can be used to fine-tune the large pre-trained model or the data can be used to better simulate the ID distribution, facilitating the differentiation of OOD data.
NPOS ~\citep{tao2023non} proposes a non-parametric outlier synthesis technique to distinguish ID and OOD data by fine-tuning CLIP with complete ID data.
In contrast, TOE ~\citep{park2023powerfulness}, while also using CE loss to constrain the model during fine-tuning, builds on the ideas of OE by focusing on textual outliers within the CLIP framework to control the model's recognition capabilities, which differs significantly from directly using OOD images.

\textbf{PT-OOD Detection.} 
``PT-OO'' samples are OOD samples with overlap in pretraining data.
After investigating and elucidating 
 the effects of various pre-training methodologies (supervised, self-supervised) on PT-OOD detection, 
~\citet{miyai2023can} observe the low linear separability in feature space significantly degrades the PT-OOD detection performance.
They suggest using distinctive features for each instance to distinguish between ID and OOD samples.

\section{Evaluation and Application}\label{evaluationandapplication}

\setlength{\tabcolsep}{5pt}
\begin{table*}[]
\caption{Summary of Datasets. The CARLA System is a simulation platform designed for evaluating OOD Detection in the field of autonomous driving, hence its entire row is filled with dashes. 
``*''indicates that this dataset is divided by category, with some data used as ID and other data as OOD. ``/'' symbol in the Usage column indicates that this dataset can be used as both ID and OOD data.
}
\begin{tabularx}{\textwidth}{lllllll}
\hline \hline 
TASK                                     & Dataset Name         & \multicolumn{1}{l}{Data Type}                 & \# Classes & \# Samples      & Reference    & Usage \\ \hline
\multirow{8}{*}{Image   Classification}  & CIFAR-10             & \multicolumn{1}{l}{Images}                    & 10                & 60,000                   &  ~\citep{wang2023clipn, esmaeilpour2022zeroZOC}      & ID/OOD \\ \cline{2-7} 
                                         & CIFAR-100            & \multicolumn{1}{l}{Images}                    & 100               & 60,000                   &  ~\citep{wang2023clipn, esmaeilpour2022zeroZOC}      & ID/OOD \\ \cline{2-7} 
                                         & MNIST                & \multicolumn{1}{l}{Images}                    & 10                & 70,000                   &  ~\citep{hendrycks2016baseline, zhao2024towards}   & ID/OOD \\ \cline{2-7} 
                                         & ImageNet-1K          & \multicolumn{1}{l}{Images}                    & 1,000             & 1,431,167                &  ~\citep{ming2022delvingMCM, wang2023clipn}     & ID \\ \cline{2-7} 
                                         & iNaturalist          & \multicolumn{1}{l}{Images}                    & 5,089             & 675,170                  &  ~\citep{ming2022delvingMCM, wang2023clipn}       & OOD \\ \cline{2-7} 
                                         & SUN                  & \multicolumn{1}{l}{Images}                    & 397               & 108,754                  &  ~\citep{ming2022delvingMCM, wang2023clipn}       & OOD \\ \cline{2-7} 
                                         & Places               & \multicolumn{1}{l}{Images}                    & \textgreater{}205 & \textgreater{}2,500,000  &  ~\citep{ming2022delvingMCM, wang2023clipn}        & OOD \\ \cline{2-7} 
                                         & Textures             & \multicolumn{1}{l}{Images}                    & 47                & 5,640                    &  ~\citep{ming2022delvingMCM, wang2023clipn}      & OOD \\ \hline
\multirow{2}{*}{Semantic  Segmentation}  & Cityscapes           & \multicolumn{1}{l}{Images}                    & 30                & 25,000                   &  ~\citep{gao2023atta}       & ID \\ \cline{2-7} 
                                         & Road Anomaly Dataset & \multicolumn{1}{l}{Images}                    & \textgreater{} 5                 & 60                       &  ~\citep{gao2023atta}       & OOD \\ \hline
\multirow{4}{*}{Object Detection}        & PASCAL VOC           & \multicolumn{1}{l}{Images}                    & 20                & 2,913                    &  ~\citep{nguyen2024zero, du2022siren}    & ID \\ \cline{2-7} 
                                         & BDD100K              & \multicolumn{1}{l}{Images}                    & 10                & 100,000                  &  ~\citep{nguyen2024zero, du2022siren}    & ID \\ \cline{2-7} 
                                         & MS-COCO              & \multicolumn{1}{l}{Images}                    & 80                & \textgreater{} 930       &  ~\citep{nguyen2024zero, du2022siren}      & OOD   \\ \cline{2-7}
                                         & OpenImages           & \multicolumn{1}{l}{Images}                    & 601               & 1761                     &  ~\citep{nguyen2024zero, du2022siren}      & OOD    \\ \hline
Autonomous Driving                       & CARLA System         & \multicolumn{1}{l}{-}                         & -                 & -                        &  ~\citep{mao2024language}  & - \\ \hline
Medical Image Analysis                   & Kvasir-Capsule*      & \multicolumn{1}{l}{Images}                    & 14                & 4,741,621                &  ~\citep{tan2024endoood}    & ID/OOD \\ \hline
\multirow{2}{*}{Text Category}           & News Category*       & \multicolumn{1}{l}{Text}                      & 41                & 210,000                  &  ~\citep{arora2021types}     & ID/OOD \\ \cline{2-7} 
                                         & SST-2                & \multicolumn{1}{l}{Text}                      & 2                 & 215,154                  &  ~\citep{arora2021types}      & ID/OOD \\ \hline
\multirow{3}{*}{Intent Detection}        & CLINC150*            & \multicolumn{1}{l}{Text}                      & 150               & 22,500                   &  ~\citep{zhan2022closer}     & ID/OOD \\ \cline{2-7} 
                                         & Banking77*           & \multicolumn{1}{l}{Text}                      & 77                & 13,083                   &  ~\citep{zhan2022closer}    & ID/OOD \\ \cline{2-7} 
                                         & StackOverflow*       & \multicolumn{1}{l}{Text}                      & 20                & 20,000                   &  ~\citep{zhan2022closer}      & ID/OOD \\ \hline
\multirow{2}{*}{Audio}                   & MSCW                 & \multicolumn{1}{l}{Audio}                     & 31                & \textgreater{}23,400,000 &  ~\citep{bukhsh2023out}      & ID/OOD \\ \cline{2-7} 
                                         & Vocalsound           & \multicolumn{1}{l}{Audio}                     & 6                 & 21,024                   &  ~\citep{bukhsh2023out}        & ID/OOD \\ \hline
Graph data                               & TU/OGB*              & \multicolumn{1}{l}{Graph data}                & 10                & 19,766                   &  ~\citep{wang2024goodat}    & ID/OOD \\ \cline{2-7} 
\hline \hline 
\end{tabularx}
\label{tab:datasets}
\end{table*}

\subsection{Evaluation metrics} 
In the vast majority of OOD detection tasks in the visual domain, the following evaluation metrics are commonly used:

\textbf{AUROC (Area Under the Receiver Operating Characteristic curve)}. 
This metric quantifies the likelihood that a classifier will assign higher scores to ID samples compared to OOD samples.
An elevated AUROC value is indicative of superior model performance, signifying an enhanced ability to distinguish between ID and OOD instances. 
Consequently, a higher value is desirable.

\textbf{AUPR (Area under the Precision-Recall curve)}. 
This metric is pertinent when the ID class is considered the positive class and is particularly valuable in the context of imbalanced class distributions. 
It assesses the balance between precision and recall, with a higher AUPR value indicating superior model performance.

\textbf{FPR@95 (False Positive Rate at 95\% True Positive Rate)}. 
This metric delineates the false positive rate (FPR) at the juncture where the true positive rate (TPR) reaches 95\%. 
It essentially gauges the proportion of OOD samples erroneously identified as ID, thus providing insight into the model's propensity for false alarms at a high sensitivity threshold. 
A reduced FPR@95\% TPR is indicative of a model's enhanced specificity in correctly flagging OOD samples while maintaining high sensitivity towards ID samples. 
Therefore, a lower value is desirable.




\subsection{Experimental Protocols}

In the traditional experimental protocol for OOD detection, test data is exclusively classified as either ID or OOD. However, as the field has advanced, there is now a more nuanced distinction between OOD and ID data, which has led to variations in the evaluation process.

Subsequently, OOD data is categorized into near-OOD and far-OOD based on the degree of covariate shift from ID data.
This categorization corresponds to dividing OOD detection tasks into near-OOD and far-OOD detection.
It is evident the near-OOD detection task is more challenging, 
however, numerous methods ~\citep{fort2021exploring, ming2022delvingMCM, bai2023id} have demonstrated excellent performance in this area.

Recently, ~\citet{yang2023full_food, bai2023feed} propose that we should consider cases where covariate shift occurs in ID data, which is not taken into account previously.
This is crucial to prevent the loss of model generalization. 
The samples mentioned earlier are termed as cs-ID data, an abbreviation for ``covariate shift ID'' data. 
Consequently, a new experimental protocol has been explored, called full-spectrum OOD detection.
During the testing phase, the model is expected to identify near-OOD and far-OOD instances. Additionally, it should refuse to provide predictions for the OOD data and accurately predict ID and cs-ID data.

\subsection{Application}
This section presents a comprehensive review of both academic and real-world applications of OOD detection. To promote further research and enable standardized evaluation, commonly used benchmark datasets are systematically summarized in Table~\ref{tab:datasets}. 

To comprehensively review recent advances in OOD detection, we first categorize our discussion by key technical domains (\textit{Computer Vision}, \textit{Natural Language Processing}, and \textit{Beyond}), as each has distinct methods and theoretical bases shaping OOD approaches. This structure enables a systematic analysis of core innovations and challenges in each field. We then adopt an application-oriented perspective (see Table~\ref{tab:application}), highlighting how these techniques are adapted and implemented in real-world scenarios.

\begin{table}[htbp]
\centering
\caption{Representative applications of OOD detection across five major domains, highlighting key use cases in industrial, medical, safety, scientific, and general-purpose settings.}
\label{tab:ood_applications}
\resizebox{0.85\textwidth}{!}{
\begin{tabular}{>{\centering\arraybackslash}p{3cm}|p{3.5cm}|p{7.2cm}|p{1.5cm}}
\hline\hline
\textbf{Domain} & \textbf{Task} & \textbf{Application Description} & \textbf{Reference} \\
\hline

\multirow{6}{3cm}{\centering Industrial \\Applications}
    & Autonomous Driving & Detecting unseen obstacles or traffic scenarios to ensure driving safety in autonomous vehicles. & \cite{tesla} \\
    \cline{2-4}
    & 3D Object Detection & Identifying OOD objects in 3D industrial environments, such as warehouses or robotics. & \cite{3D,li5256108out} \\
    \cline{2-4}
    & Object Detection & Finding anomalous or unknown items in industrial production lines. & \cite{syed2024situation,chen2024protoood,zhang2025vista} \\
    \cline{2-4}
    & Earth Observation Imagery & Detecting novel land-cover types or changes in satellite and remote sensing images. & \cite{lebellier2024detecting} \\
    \cline{2-4}
    & Edge Computing & Real-time detection of abnormal data on edge devices to enhance responsiveness and security. & \cite{li2025secure} \\
    \cline{2-4}
    & Time Series Analysis & Monitoring equipment by identifying abnormal patterns for predictive maintenance. & \cite{website_key} \\
\hline

\multirow{2}{3cm}[-3pt]{\centering Healthcare \\and \\Biology}
    & Medical Image Analysis & Detecting unknown diseases or anomalies in medical images (e.g., X-ray, MRI). & \cite{Diseases,pokhrel2024tta,smedsrud2021kvasir} \\
    \cline{2-4}
    & Biological Classification & Identifying previously unseen species and enhancing biodiversity monitoring. & \cite{arslan2024talics3} \\
\hline

\multirow{4}{3cm}{\centering Security\\and\\Surveillance}
    & Facial Authentication & Detecting spoofed or unregistered faces for enhanced security. & \cite{kahya2024food} \\
    \cline{2-4}
    & Facial Expression Recognition & Identifying unknown or abnormal facial expressions in surveillance. & \cite{zhang2024open} \\
    \cline{2-4}
    & Human Action Recognition & Detecting anomalous or dangerous actions in monitored environments. & \cite{sim2023simple,mandal2019out,xu2024skeleton} \\
    \cline{2-4}
    & Audio Recognition & Recognizing abnormal sounds such as alarms or intrusions. & \cite{mazumder2021multilingual} \\
\hline

\multirow{5}{3cm}{\centering Scientific\\Research}
    & Astronomical Imaging & Discovering unknown celestial objects or phenomena in astronomical data. & \cite{zhou2024evaluating} \\
    \cline{2-4}
    & Solar Image Analysis & Detecting novel solar activities or anomalies in solar observation images. & \cite{das2024interpretable} \\
    \cline{2-4}
    & Graph Neural Networks & Detecting anomalies in complex scientific data such as molecular structures or social networks. & \cite{bazhenov2022towards,wu2023energy} \\
    \cline{2-4}
    & Mathematical Reasoning & Identifying OOD problem types in automated reasoning systems. & \cite{wang2024embedding} \\
    \cline{2-4}
    & Spiking Neural Networks & Recognizing abnormal neural signals in bio-inspired computing models. & \cite{terres5152073forward,orchard2015converting} \\
\hline

\multirow{7}{3cm}{\centering General-Purpose\\Applications}
    & Image Classification & Detecting unknown categories in image classification systems. & \cite{yang2023auto, yang2022out, sun2021react} \\
    \cline{2-4}
    & Semantic Segmentation & Identifying unknown regions or objects in pixel-level segmentation tasks. & \cite{gao2023atta, lis2019detecting} \\
    \cline{2-4}
    & Text Classification & Detecting unseen topics or intents in text data. & \cite{zhan2024vi} \\
    \cline{2-4}
    & Intent Detection & Recognizing new user intents in intelligent assistants and chatbots. & \cite{wang2024beyond, zhan2022closer} \\
    \cline{2-4}
    & Question-Answering Systems & Detecting questions that are out-of-scope or unanswerable by the system. & \cite{zhang2024your} \\
    \cline{2-4}
    & Document Classification & Detecting anomalies in documents containing multiple data modalities. & \cite{constantinou2024out} \\
    \cline{2-4}
    & Reinforcement Learning & Detecting abnormal behavior or novel states in agent-environment interactions. & \cite{chen2025taming,sedlmeier2019uncertainty} \\
\hline\hline
\end{tabular}
\label{tab:application}
}
\end{table}

\subsubsection{Computer Vision}

Most of the efforts in OOD detection have been devoted to the field of computer vision, we list extensive vision-related tasks as follows:
\begin{itemize}
\item \textbf{Image Classification.}
Image classification represents the primary testbed for OOD detection research. Common ID datasets include MNIST~\citep{xiao2017fashion}, CIFAR~\citep{krizhevsky2009learning}, and ImageNet-1K~\citep{deng2009imagenet}, while OOD evaluation typically uses datasets like iNaturalist~\citep{van2018inaturalist}, SUN~\citep{xiao2010sun}, and Textures~\citep{cimpoi14describing}. Recent research categorizes OOD datasets based on their detection difficulty: near-OOD (e.g., SSB-hard~\citep{vaze2021open}, NINCO~\citep{bitterwolf2023or}) for harder-to-distinguish categories, and far-OOD (e.g., iNaturalist, OpenImage-O~\citep{wang2022vim}) for more distinct categories.

    \item \textbf{Semantic Segmentation.} 
    Recent works ~\citep{gao2023atta} have started delving into the dense OOD detection task, also known as anomaly segmentation. 
    The datasets used for evaluation include the Cityscapes dataset ~\citep{cordts2016cityscapes}, the Road Anomaly dataset ~\citep{lis2019detecting}, and the recently developed SOOD-ImageNet ~\citep{bacchin2024sood}.

    \textbf{Object Detection.}  
OOD detection for object detection tasks is a relatively recent research area~\citep{du2022siren}, with evaluations commonly conducted on datasets such as PASCAL-VOC~\citep{Everingham15} and BDD-100K~\citep{yu2018bdd100k}. Recent methods explore diverse strategies, including ensemble architectures~\citep{syed2024situation}, prototype-based similarity~\citep{chen2024protoood}, and visual-contextual augmentation~\citep{zhang2025vista}.

    \item \textbf{3D Object Detection.} 
    OOD detection in LiDAR-based 3D object detection generates synthetic OOD data and trains MLPs to distinguish ID and OOD targets, with new evaluation protocols for realistic scenarios~\citep{kosel2024revisiting}.

    \item \textbf{Autonomous Driving.} 
    Autonomous driving is a critical application of OOD detection. Recent work uses the CARLA simulator~\citep{dosovitskiy2017carla} to evaluate OOD detection performance in driving scenarios~\citep{mao2024language}.  ~\citet{mao2024language} propose a language-augmented latent representation method, leveraging the image-text cosine similarity from the CLIP model to improve transparency and controllability in detection. Another recent work focus on real-time OOD perception in trajectory prediction, further enhancing vehicle safety~\citep{banerjee2024building}.
    In real-world applications, Tesla’s “Shadow Mode”~\cite{tesla} identifies cases where AI predictions differ from human drivers in real traffic—these OOD samples are collected and used to retrain and improve autonomous driving models.

    \item \textbf{Medical Image Analysis.} 
    OOD detection is vital in medical imaging, using datasets like CIFAR-10 and Kvasir-Capsul~\citep{smedsrud2021kvasir} depending on the image category~\citep{tan2024endoood}. Recent studies focus on reliable OOD detection methods for digital pathologyfor improving OOD detection in gastrointestinal vision~\citep{pokhrel2024tta}.
    In practice, IDx-DR~\citep{Diseases}, the first FDA-approved autonomous AI diagnostic system, implements OOD detection by rejecting images that fail to meet quality standards or contain anomalies, deferring these cases to human experts for assessment.
   \item \textbf{Human Action Recognition.} 
    OOD detection plays a crucial role in ensuring the robustness of Human Action Recognition (HAR) models, particularly when faced with previously unseen actions. By leveraging advanced techniques such as attention-based debiasing~\citep{sim2023simple}, generative feature synthesis~\citep{mandal2019out} and energy-based skeleton modeling~\citep{xu2024skeleton}, recent research has significantly improved OOD detection performance.
    
    \item \textbf{Solar Image Analysis.}
    In space weather forecasting, OOD detection serves as a crucial tool for identifying solar anomalies. Recent unsupervised ~\citep{ das2024interpretable}approaches leverage Solar Dynamics Observatory data to enhance detection accuracy. 
    \item \textbf{Facial Expression Recognition.} 
    In facial expression analysis, the challenge of distinguishing unknown expressions from known categories has led to advanced OOD detection methods. Recent approach~\citep{zhang2024open} focuses on leveraging the unique characteristics of facial expression features, incorporating attention map consistency and cycle training mechanisms to effectively identify OOD samples.
    \item \textbf{Facial Authentication.} 
    In facial authentication systems, OOD detection is increasingly important for ensuring robust and secure identity verification. The FOOD framework~\citep{kahya2024food} exemplifies this trend by employing 60 GHz FMCW radar and a convolutional encoder-decoder architecture to achieve both accurate classification and reliable anomaly detection.
    \item \textbf{Earth Observation Imagery.} 
    For earth observation imagery, OOD detection enhances the identification of rare or unexpected events, such as natural disasters. Diffusion models like ODEED~\citep{lebellier2024detecting} have demonstrated strong performance in detecting OOD samples, with evaluations conducted on benchmark datasets such as SpaceNet 8 to validate their effectiveness.

\end{itemize}

\subsubsection{Natural Language Processing}
OOD detection has become a crucial research topic in NLP, especially with the advent of LLMs and their deployment in real-world applications~\citep{liu2025survey, lang2023survey_nlpood}. Below, we summarize key tasks and representative methods:

\vspace{-3pt} 
\begin{itemize}
    \item \textbf{Intent Detection.} 
    Intent Detection is a significant application of OOD detection in NLP, particularly in dialogue systems, where identifying user intentions that fall outside the predefined set of intents is crucial~\citep{wang2024beyond, zhan2022closer}. Common datasets for evaluation include CLINC150~\citep{larson2019evaluation}  and HWU64~\citep{liu2020benchmarking}. These datasets are specifically designed to evaluate OOD detection in intent classification tasks.

    \item \textbf{Text Classification.} 
    In text classification tasks, OOD detection helps identify texts that do not belong to any of the known categories. Datasets such as News Category~\citep{misra2022news} and SST-2~\citep{socher2013recursive} are frequently used for ID data, with 20 Newsgroups (20NG)~\citep{lang1995newsweeder} often serving as an OOD dataset. Methods like VI-OOD~\citep{zhan2024vi} enhance detection by optimizing joint distributions.

    \item \textbf{Question-Answering Systems.} Question-answering systems require robust mechanisms to handle diverse user queries while maintaining response accuracy. The challenge of identifying questions that fall outside the system’s knowledge domain has led to sophisticated OOD detection approaches.Recent LLM-based methods~\citep{zhang2024your} leverage likelihood ratios between pretrained and fine-tuned models for effective OOD detection.

    \item \textbf{Mathematical Reasoning.} Mathematical problem-solving systems face challenges when encountering problem types not seen during training. Using MultiArith~\citep{roy2016solving} as the primary ID dataset, researchers evaluate OOD detection against diverse benchmarks. Recent advances, such as TV-Score~\citep{wang2024embedding}, analyze embedding trajectory variations to enhance OOD detection performance.
    
    \item \textbf{Document Classification}. In (multimodal) document classification, where documents integrate both text and images, OOD detection is essential for identifying unseen document types. Datasets such as Tobacco3482~\citep{aggarwal2017tobacco} and FinanceDocs~\citep{constantinou2024out} are commonly used for evaluation. The AHM~\citep{constantinou2024out} approach enhances OOD detection by improving the representation of multimodal features.

\end{itemize}

\subsubsection{Beyond Computer Vision and Natural Language Processing}
In addition to the two data modalities mentioned above, OOD detection still has many important applications across various types of data.

\vspace{-3pt} 

\begin{itemize}
  
\vspace{-1pt}
\item \textbf{Audio Recognition.} OOD detection in audio recognition has evolved from traditional probabilistic scoring~\citep{lane2006out} to methods leveraging confidence-based classification~\citep{lane2006out}, neural embeddings~\citep{ryu2017neural}, and deep nearest neighbor approaches~\citep{bukhsh2023out}. Recent advances include autoencoder models based on WavLM~\citep{du2024towards}, which are particularly effective for synthetic speech detection. Standard benchmarks such as MSCW~\citep{mazumder2021multilingual} support comprehensive evaluation in this domain.

\item \textbf{Graph Neural Networks.} In GNNs, OOD detection encompasses node- and graph-level tasks, utilizing uncertainty estimation~\citep{bazhenov2022towards}, energy-based methods~\citep{wu2023energy}, and semi-supervised learning~\citep{huang2022end}. Frameworks like HGOE~\citep{junwei2024hgoe} and PGR-MOOD~\citep{shen2024optimizing} introduce anomalous or pseudo-OOD samples to enhance detection. GOLD~\citep{wang2025gold} further advances this area by generating pseudo-OOD data through latent generative modeling. Evaluation commonly relies on datasets from TU~\citep{morris2020tudataset_TU} and OGB~\citep{hu2020open_OGB}.

    \item \textbf{Reinforcement Learning.} OOD detection is vital in reinforcement learning for improving robustness to novel states and actions~\citep{leonardos2023addressing}. Approaches typically involve uncertainty quantification~\citep{sedlmeier2019uncertainty}, environment modifications~\citep{mohammed2021benchmark}, and offline RL techniques such as ADAC~\citep{chen2025taming}. Evaluation frameworks often focus on environmental perturbations~\citep{mohammed2021benchmark} to benchmark robustness.
    \item \textbf{Time Series Analysis.} Time series OOD detection faces unique challenges due to temporal dependencies. Recent research explores modality-agnostic approaches for multivariate time series~\citep{gungor2025ts}, with practical applications in fraud detection systems like AWS Fraud Detector~\citep{website_key}.
    \item \textbf{Biological Classification.} OOD detection identifies unseen species in biological classification. DNA barcodes enhance image-based classification through re-ranking, improving detection accuracy~\citep{arslan2024talics3}.
    \item \textbf{Spiking Neural Networks (SNN).} SNNs’ sparse representations challenge OOD detection. Recent methods combine feedforward learning with distance-based scoring~\citep{terres5152073forward}. The N-MNIST dataset~\citep{orchard2015converting} is commonly used for evaluation, with OOD samples generated through diverse event patterns.
    \item \textbf{Astronomical Imaging.} OOD detection distinguishes between simulated and real galaxy images. Bayesian comparison and sparse variational autoencoders~\citep{zhou2024evaluating} uncover subtle differences.
    \item \textbf{Edge Computing.} Edge computing requires efficient, privacy-preserving OOD detection. SecDOOD~\citep{li2025secure} uses cloud-device architecture and HyperNetwork generation to avoid device-side backpropagation. CIFAR-10 and Tiny ImageNet serve as evaluation datasets.
\end{itemize}

The deployment of OOD detection in high-stakes applications introduces considerations beyond predictive accuracy, spanning regulatory compliance, operational safety, and algorithmic fairness. 
For compliance, OOD detection supports safety standards like ISO~21448\footnote{https://www.iso.org/standard/77490.html} by identifying OOD scenarios, informing release decisions and enabling in-operation validation via shadow-mode monitoring~\citep{henriksson2023out}. 
For safety, approaches~\citep{henriksson2023out,lin2025safety} are advancing from runtime input flagging~\citep{henriksson2023out} toward monitoring system-level properties for predictive safety assurance with formal guarantees~\citep{lin2025safety}. 
For fairness, mitigating bias against minority subgroups involves developing fairness-aware algorithms~\citep{shekhar2021fairod} and constructing unbiased benchmarks that control for semantic ambiguity~\citep{inkawhich2022improving}.

\vspace{-5pt}
\section{Emerging Trends And Open Challenges}\label{emerging future work}
Despite rapid advancements in OOD detection, numerous emerging trends and less-explored challenges remain.
In this section, we explore emerging trends and open challenges from three distinct perspectives: methodologies, scenarios, and applications.

\subsection{Better methodologies of OOD detection}
\textbf{Meta-Learning Adaptation.} Addressing the challenge of adapting to new data distributions at test time, meta-learning has gained attention for its ability to facilitate efficient OOD detection in novel scenarios. Recent work also highlights the importance of developing better sampling strategies to more effectively leverage outlier information during training~\citep{ming2022poem}.

\textbf{Model Selection.} Choosing an appropriate OOD detection method for a specific task is challenging, as performance varies across data distributions. Automated frameworks like MetaOOD~\citep{qin2024metaood} use meta-learning for zero-shot selection, while DSDE~\citep{geng2024dsde} and score-combining methods~\citep{reehorst2025score} apply proportion estimation and likelihood ratio tests to improve ensemble and score fusion. Notably, recent studies show that partially trained models can also be effective for OOD detection~\citep{montazeran2025ood}, highlighting new possibilities for leveraging model dynamics.



\textbf{Active Learning.} Active learning prioritizes labeling informative data points for efficiency in data-scarce settings. In OOD detection, it focuses on ambiguous or uncertain samples to improve ID and OOD discrimination. SISOM~\citep{schmidt2024unified} combines active learning and OOD detection using feature space distances.

\textbf{Continual and Incremental Learning.} Continual and incremental learning enable OOD detection models to adapt to streaming data while preserving previous knowledge. Recent work includes unsupervised OOD detection~\citep{doorenbos2024continual}, OpenCIL benchmark~\citep{miao2024opencil}, and hierarchical two-sample tests~\citep{liu2025h2st}.

\textbf{Theoretically-Driven Score Design.} While many post-hoc scoring methods have been proposed for OOD detection in single-modal regimes, the transition to multi-modal domains necessitates more principled, theoretically motivated score designs. For example, MCM~\citep{ming2022delvingMCM} extends softmax-based approaches, but further innovation is needed to better capture the complex relationships in multi-modal data.

\vspace{-5pt}
\subsection{More practical scenarios of OOD detection}
Under the current trend, there is a growing need for the emergence of more practical scenarios, driven by the limitations of existing impractical restrictions.

\textbf{Quick Test-Time Adaptation.} The introduction of test-time adaptation scenarios, such as CAOOD~\citep{wu2023metaMOL}, marks a step forward in making OOD detection more reliable and adaptable in dynamic real-world environments.


\textbf{Federated Environments.} Federated learning involves training models across decentralized clients, each with its own data distribution~\citep{guo2025exploring,kuang2024enhanced}. This distributional variability presents unique OOD detection challenges. Methods like FOOGD~\citep{liao2024foogd} address these issues by enhancing robustness to both covariate and semantic shifts in federated settings.

\textbf{Multi-Modal Detection.} With the increasing prevalence of multi-modal data, OOD detection has expanded beyond single modalities. MultiOOD~\citep{dong2024multiood} established a comprehensive benchmark with algorithms like A2D and NP-Mix. Recent works~\citep{gao2024no} have advanced cross-modal alignment techniques.

\textbf{Multi-Label OOD Detection.} Real-world applications often involve multi-label data, where each instance may belong to multiple categories. Detecting OOD samples in these scenarios is challenging due to increased dimensionality and label correlations. Recent approaches~\citep{mei2024multi} leverage uncertainty modeling and subjective logic to improve robustness in multi-label OOD detection.

\textbf{Open-Vocabulary Scenario.} Traditional approaches often assume access to all ID categories, an assumption that does not hold in open-vocabulary contexts. Recent work~\citep{li2024learning} explores learning migratable negative prompts, enabling OOD detection even when only a subset of ID labels is available, thus paving the way for more generalizable models.

\textbf{Noisy Label Environments.} Most prior studies focus on clean label settings, but real-world data often contains label noise. Investigations into noisy label scenarios~\citep{humblot2024noisy} offer insights and practical recommendations for improving OOD detection robustness under such conditions.


\subsection{New applications of OOD detection}

\textbf{Additional Modalities.} While significant advances have been made in vision and language domains, OOD detection in modalities such as speech and physiological signals remains underexplored. These areas, particularly emotion-related physiological signals with high inter-subject variability, present promising directions for future research~\citep{dong2024multiood}.

\textbf{Human-in-the-Loop Systems.} Human expertise integration in OOD detection is crucial for safety-critical applications. Recent frameworks~\citep{vishwakarma2024taming,bai2024out,bai2024aha} leverage human feedback to optimize detection thresholds and guide annotation.

\textbf{Web Image Scraping.} OOD detection is also finding novel applications in automating web image scraping. For example, zero-shot ID detection~\citep{miyai2023zero} classifies images as ID if they contain relevant objects, offering a fresh perspective on leveraging OOD detection for large-scale, real-world data acquisition.

\section{Conclusion}\label{conclusion}
OOD detection is a critical component for trustworthy machine learning. 
In this paper, we provide a comprehensive review of recent advances in OOD detection, focusing for the first time on the problem scenario perspective: training-driven, training-agnostic, and large pre-trained model-based OOD detection. 
We also summarize extensively used evaluation metrics, experimental protocols, and diverse applications. 
We believe that our taxonomy of existing papers and extensive discussion of emerging trends will contribute to a better understanding of the current state of research, assist practitioners in selecting suitable approaches, and inspire new research hotspots.

\section{Acknowledgements}

We thank the editor and reviewers for their constructive feedback, and Zhen Jia and Yongcan Yu for their assistance with the manuscript. This work was supported by the National Natural Science Foundation of China (62276256, U2441251), the Young Elite Scientists Sponsorship Program by CAST (2023QNRC001), and the Young Scientists Fund of the State Key Laboratory of Multimodal Artificial Intelligence Systems (ES2P100117).

\newpage
\bibliographystyle{unsrtnat}
\bibliography{main}

\end{document}